\documentclass[review,12pt,3p,number]{elsarticle}

\usepackage{amssymb}
\usepackage{amsmath}
\usepackage[inline]{enumitem}
\usepackage{microtype}
\usepackage{graphicx}
\usepackage{subcaption}
\usepackage{booktabs}
\usepackage{multirow}
\usepackage{float}
\usepackage{tabularx}
\usepackage{array}
\newcolumntype{Y}{>{\raggedright\arraybackslash}X}
\usepackage{longtable} 
\usepackage{etoolbox}
\usepackage[pagewise]{lineno}
\usepackage[hidelinks]{hyperref}
\usepackage[nameinlink]{cleveref}
\crefname{figure}{Fig.}{Figs.}
\usepackage{adjustbox}
\usepackage{caption}
\usepackage[section]{placeins}

\AtBeginEnvironment{table}{\small}
\AtBeginEnvironment{tabular}{\small}
\AtBeginEnvironment{tabularx}{\small}
\usepackage{caption}

\begin{document}

\begin{frontmatter}

\title{Real-time prediction of workplane illuminance distribution for daylight-linked controls using non-intrusive multimodal deep learning}
\author[label1]{Zulin Zhuang} 
\author[label1]{Yu Bian\corref{cor1}} 
\cortext[cor1]{Corresponding author.}
\ead{aryubian@scut.edu.cn}
\affiliation[label1]{organization={School of Architecture; State Key Laboratory of Subtropical Building and Urban Science, South China University of Technology},
            city={Guangzhou},
            state={Guangdong},
            country={China}}
\begin{abstract}
Daylight-linked controls (DLCs) have significant potential for energy savings in buildings, especially when abundant daylight is available and indoor workplane illuminance can be accurately predicted in real time. Most existing studies on indoor daylight predictions were developed and tested for static scenes. This study proposes a multimodal deep learning framework that predicts indoor workplane illuminance distributions in real time from non-intrusive images with temporal-spatial features. By extracting image features only from the side-lit window areas rather than interior pixels, the approach remains applicable in dynamically occupied indoor spaces. A field experiment was conducted in a test room in Guangzhou (China), where 17,344 samples were collected for model training and validation. The model achieved $R^2 > 0.98$ with RMSE $< 0.14$ on the same-distribution test set and $R^2 > 0.82$ with RMSE $< 0.17$ on an unseen-day test set, indicating high accuracy and acceptable temporal generalization.
\end{abstract}

\begin{highlights}
\item Extract features from the window area in images instead of interior pixels.
\item Fuse images with temporal and spatial features for accurate daylight prediction.
\item Validate the model's temporal generalization on data from a new day.
\end{highlights}


\end{frontmatter}

\section{Introduction}
\label{sec1}
Global warming and the energy crisis have driven increasing attention to energy efficiency and carbon emissions reductions \cite{wang2020marginal}. Among various sectors, buildings are a major contributor to both energy consumption and carbon emissions \cite{chen2022strategies}. Building operations are responsible for about 30\% of global energy consumption and 26\% of energy-related CO$_2$ emissions \cite{iea_buildings2025}. With buildings, lighting systems are one of the largest electricity consumers \cite{nicol2006using} and account for 15\% to 45\% of building electricity use in operation \cite{norouziasl2023identifying}. Effective lighting control strategies have proven in reducing energy use \cite{bellia2016daylight, williams2012lighting}. There are three kinds of control system, occupancy-based controls, time-based controls and daylight-linked controls (DLCs) \cite{ul2014review}. Among these, DLCs, which regulate electric lighting in response to daylight availability, have the greatest potential for energy savings while also ensuring occupants' visual comfort, and are therefore recommended for implementation in building operations \cite{li2003analysis, bellia2024testing, gentile2022evaluation}.

Although DLCs are widely recognized as an effective energy-saving strategy, their practical implementation remains limited due to several factors \cite{ul2014review}. One key barrier is the difficulty of accurately predicting real-time indoor daylight environment during both design and operation of DLCs \cite{bellia2016daylight, park2003workplane}. DLCs typically control constant illuminance levels rely on convert signals from photosensors or images \cite{kim2022performance, wagiman2020lighting}. These sensing devices are usually non-intrusively deployed on ceilings or walls to avoid obstructing tasks or being shaded \cite{do2023selection, doulos2019minimizing}. Consequently, workplane illuminance must be inferred indirectly from these sensor signals rather than measured directly, making additional system uncertainty \cite{park2003workplane}. If one-to-one mapping between sensor signals and workplane illuminance could be established, perfect control would be achievable \cite{lee1999effect}. However, inaccurate predictions can lead to improper dimming responses in DLCs, resulting in either excessive or insufficient illuminance, thereby affecting both the energy efficiency and visual comfort \cite{li2010analysis, bonomolo2017set}. In this context, there is a critical need for more reliable methods to predict indoor daylight illuminance distribution in real time, to support effective implementation of DLCs. 

Previous studies have explored various approaches to address this prediction challenge from multiple aspects, mainly including sensor-based mapping, imaging techniques, and data-driven methods \cite{bellia2016daylight, liu2016fuzzy}. Sensor-based mapping methods calibrate photosensor signals to infer workplane illuminance \cite{choi2005characteristics}. This method needs to address issues including appropriate photosensor placement and photosensor spatial sensitivity curve selection \cite{bonomolo2017set, bellia2015lighting}, and often relies on multiple sensors that would cause data overload \cite{li2014design, abbas2015survey}. Do et al.\cite{do2023selection} evaluated diverse ceiling photosensors from full cosine to 15cos and placements based on annual daylight simulations. Imaging techniques use calibrated charge-coupled device (CCD) or complementary metal-oxide semiconductor (CMOS) cameras to acquire luminance maps and convert them into spatial illuminance distributions using algorithms \cite{sarkar2006novel, newsham2009camera}. For daylight harvesting, this method exceeds the performance of current systems, as a single camera can replace multiple sensors by capturing the full luminance distribution \cite{newsham2009camera, kruisselbrink2018photometric}. Sarkar \cite{sarkar2008integrated}, Bellia \cite{bellia2011illuminance}, Moeck \cite{moeck2006illuminance} , Mead \cite{mead2017ubiquitous} and Kamath \cite{kamath2022prediction} have adopted this method to extract illuminance measurements. Data-driven methods use machine learning models to predict workplane illuminance distributions from various input features \cite{colaco2012integrated, ayoub2020review}. These models learn the complex nonlinear relationships between inputs (like time, sky conditions, window properties, etc.) and illuminance outputs \cite{jordan2015machine}, offering a faster alternative to conventional simulation-based methods \cite{ngarambe2020comparative}. Beccali et al. \cite{beccali2018assessment} used artificial neural networks (ANNs) to optimize sensor placement for improved prediction accuracy.

However, it is important to note that fluctuations of daylight can cause notable uncertainties in evaluating daylight availability \cite{kubba2017components}. While existing studies have proposed various methods for predicting indoor daylight distributions, the challenge lies in how to ensure accurate and real-time illuminance predictions under dynamically changing scenarios \cite{bellia2016daylight}. Factors such as changing weather and sun position can lead to rapid temporal fluctuations in indoor daylight environment, thereby resulting in spatially uneven distributions across indoor areas \cite{park2013proposal, lee2024lighting}. Lou et al. \cite{lou2024investigation} quantified the duration and probability of electric lighting operation caused by daylight fluctuations under time-delay and dead-band lighting control settings. Kim and Kim \cite{kim2007impact} investigated the influence of daylight variations on dimming systems and found that partial shielding of photosensors can reduce the fluctuation of electric light output. Bellia et al. \cite{bellia2018daylight} simulated different switching systems to analysis the impact of daylight fluctuations on controls, suggesting that shading systems can reduce the effects of the daylight fluctuations on the photosensor detections. 

Recently, data-driven methods, particularly deep learning techniques, which can efficiently learn the dynamic, non-linear variations in daylight, have been suggested as an alternative to solve illuminance prediction problems \cite{ayoub2020review, belany2021combination, wagiman2020lighting}. Kandasamy et al. \cite{kandasamy2018smart} introduced an ANN-based controller for personalized lighting and daylight harvesting. Oh et al. \cite{oh2025method} proposed a deep neural network (DNN) model capable of real-time illuminance mapping using a single sensor and the sun position. Li et al. \cite{li2024predictive} developed a multimodal generative adversarial network (GAN) model that combines image and vector inputs for building feature representation, thereby improving the accuracy and generalizability of daylight related information. Besides, multimodal models offer great potential for daylight prediction by combining structured and image data for enhanced feature representation due to the significant abilities of feature representation \cite{zhang2019multimodal}. Li et al. developed a multimodal generative adversarial network (GAN) model for daylight prediction \cite{li2024predictive} and glare evaluation \cite{li2025highly}, thereby improving the accuracy and generalizability of daylight related information. Yan et al. \cite{yan2025multimodal} proposed a graph-structured feature extraction and fusion framework that integrates floorplan images, building parameters and weather conditions to predict daylight performance of residential buildings, demonstrating improved accuracy and generalizability. These studies demonstrate the potential of deep learning and multimodal models to estimate real-time daylight environment by capturing complex relationships among multiple features. However, the finite generation logic often restricts model generalizability to new scenarios \cite{han2021developing, ngarambe2022review}, and the reliance on simulated data may limit their real-world applications \cite{ayoub2020review}.

Furthermore, prediction accuracy is easily affected by dynamic indoor activities such as frequent occupant activities and furniture movements, even for non-intrusive sensing. In practice, imaging techniques are limited to static scenes, as motion during multi-exposure HDR fusion produces ghosting artifacts \cite{kalantari2017deep, xu2025adaptiveae, wang2021deep}, while changes in surface color or reflectance caused by such activities compromise luminance measurements and subsequent illuminance conversion \cite{moeck2006illuminance, guha2023review}. Liu et al. \cite{liu2024interior} proposed a model to learn the relationship among luminaire dimming levels, occupant position, and workplane illuminance, enabling real-time free-point prediction with high accuracy. Shen et al. \cite{shen2022smart} fused monocular depth estimation with multi-object detection to control the lights by identifying the distribution of personnel from the video. Mah and Tzempelikos \cite{mah2024real} developed a convolutional neural network (CNN)-based model trained with HDR images, in which a region-of-interest (ROI) program was used to predefine and label target areas including occupancy, equipment, lighting, and window in captured office images. However, these studies rely on intermediate recognition processes, such as detecting occupants or objects, where recognition errors can be accumulated and degrade the final prediction accuracy.

Existing studies have laid a solid foundation in the fields of daylight prediction and non-intrusive measurement. However, there remains a gap in the literature regarding how to accurately predict the indoor workplane illuminance distribution in real time under dynamic activity scenarios. There are several challenges that require further exploration:
(i) Real-time prediction of the indoor daylight environment usually involves multiple feature types, including weather conditions, time-related features and building conditions. These features vary in dimensions. A new method is therefore needed for integrating multiple features to improve prediction accuracy. (ii) Traditional luminance-based methods are limited to static scenarios. While existing approaches estimate illuminance by monitoring activities in target regions, prone to cumulative errors and high computational costs. (iii) Most studies generate datasets using simulation tools, which may limit their applicability due to the inability to fully capture the complexity and variability of real-world environments. Besides, HDR imaging is computationally demanding and costly to deploy in practice. To address these challenges, the contributions of this study are presented as followed:
\begin{enumerate}[label=(\arabic*)]
  \item A multimodal deep learning model is proposed to integrate diverse-dimensional data, including image data, structured data such as temporal, and spatial features, thereby improving illuminance prediction accuracy and providing a robust foundation for future studies in intelligent lighting control.
  \item A real-time indoor illuminance distribution prediction method is developed using non-intrusive images that capture daylighting features from the side-lit window region rather than the user’s workspace. Compared with previous non-intrusive approaches, this method ensures more reliable results and greater applicability in dynamic environments, as it is less affected by indoor activities.
  \item Field measurement data are adopted for model training to overcome the limitations of simulation-generated datasets. An automatic data collection process is developed to continuously monitor and record high-frequency measurements. Moreover, low-dynamic-range (LDR) images in JPEG format are employed as model inputs, which greatly simplifies the workflow, reduces computational cost, and enhances practicality for real-world applications.
\end{enumerate}
  
\section{Methodology}
\label{sec2}
\subsection{Overview workflow}
\label{subsec21}
\begin{figure}[htbp!]
  \centering
  \includegraphics[width=\linewidth,keepaspectratio]{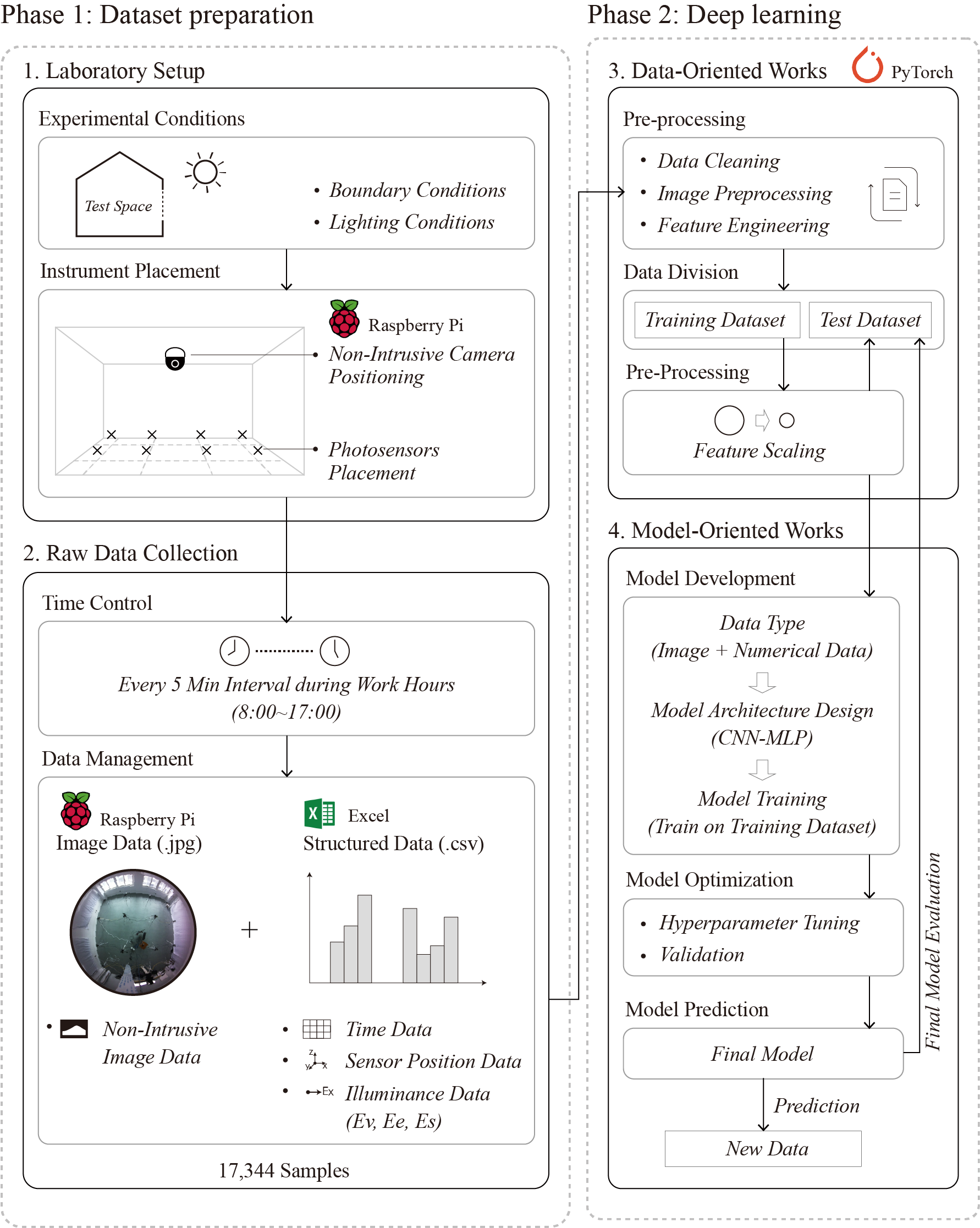}
  \caption{Overview of the workflow.}
  \label{fig:workflow}
\end{figure}

This study presents a systematic framework to predict the indoor daylighting environment built upon non-intrusive images and deep learning techniques, and validates its implementation through a field test case. The workflow, as illustrated in \cref{fig:workflow}, consists of two main phases: dataset preparation and deep learning. The objective of the first phase is to gather and organize a robust dataset to support subsequent deep learning model development. A controlled laboratory environment is established with well-defined boundary and lighting conditions. Instruments used for the experiment are positioned, with a Raspberry Pi-controlled camera mounted on the ceiling to capture non-intrusive image data, and three photosensors oriented in different directions installed at each of 16 evenly distributed measurement points. Experimental data are collected at 5-minute intervals during working hours (from 8 a.m. to 5 p.m.). In total, 17,344 samples comprising image data and structured data such as timestamps, sensor positions, and illuminance measurements were acquired to serve as the raw dataset for model development.

In the second phase, the data are pre-processed through data cleaning, image preprocessing, and feature engineering and then divided into training and test datasets. Feature scaling is subsequently applied to normalize the features. Based on the multimodal nature of the data, a convolutional neural network–multilayer perceptron (CNN–MLP) model was developed to integrate image and structured features for real-time illuminance prediction. The model is trained on the training dataset, followed by optimization through hyperparameter tuning and cross-validation to enhance its reliability. After training, the model is validated on the test dataset and used to predict new data.

\subsection{Dataset preparation}
\label{subsec22}
\subsubsection{Laboratory setup}
\label{subsubsec221}
\begin{figure}[t]
  \centering
  \includegraphics[width=\linewidth,keepaspectratio]{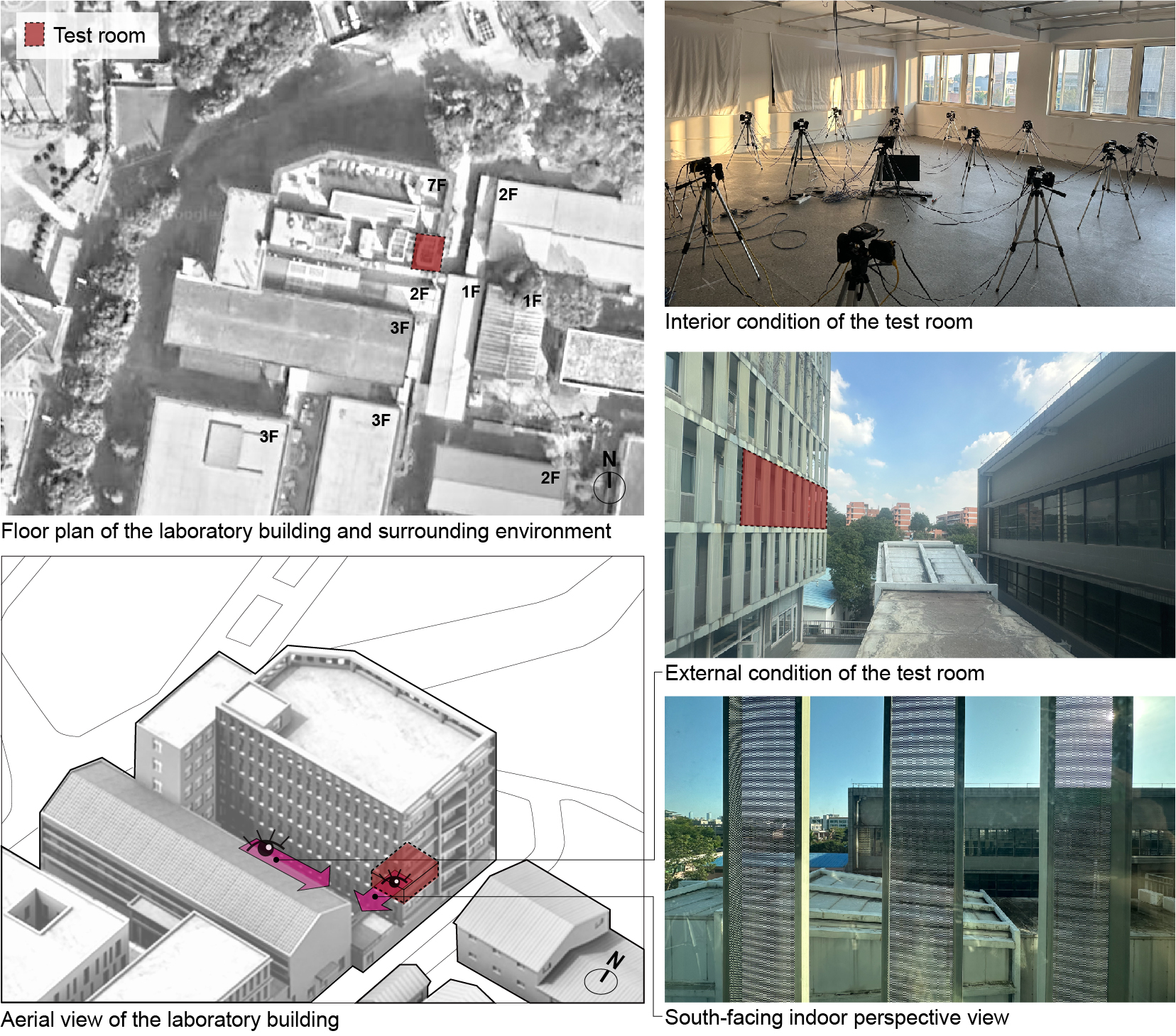}
  \caption{Boundary conditions of the test room.}
  \label{fig:BCoTR}
\end{figure}

A field study was conducted in a test room located on the third floor of a laboratory building in Guangzhou, China (Latitude: N~$23^\circ13'$, Longitude: E~$113^\circ27'$), as shown in \cref{fig:BCoTR}. The room features external horizontal louvers on its south facade, and a three-story building located approximately 9 m to the south, which may impact the room's natural lighting inside the room. The physical and environmental settings of the test room, including boundary conditions, instrument placement, and time control settings, are presented in \Cref{tab:lab_params}.

\begin{table}[htbp!]
  \centering
  
  \renewcommand{\baselinestretch}{1.0}\selectfont  
  \scriptsize                                     
  \renewcommand{\arraystretch}{1.2}                
  \setlength{\tabcolsep}{4pt}                    
  
  \caption{Laboratory settings and experimental parameters.}
  \label{tab:lab_params}
  
  \begin{tabularx}{\textwidth}{@{} l l >{\raggedright\arraybackslash}X @{}}
    \toprule
    Category & Parameter & Details \\ 
    \midrule

    \multirow[t]{6}{*}{Boundary conditions} 
      & Location                  & Guangzhou, China (Latitude: N~$23^\circ13'$, Longitude: E~$113^\circ27'$) \\
      & Room dimensions           & 7.1\,m~$\times$~8.9\,m~$\times$~2.6\,m (L–W–H) \\
      & Orientation               & $10^\circ$ east of south \\
      & South-facing windows      & 2.75\,m~$\times$~1.35\,m~$\times$~2, uncovered, closed \\
      & East-facing windows       & 2.75\,m~$\times$~1.35\,m~$\times$~2, covered with opaque panels \\
      & External obstructions     & Building 8\,m away to the south \\
      
    \cmidrule(lr){2-3} 
    
    \multirow[t]{2}{*}{Lighting conditions}
      & Daylight orientation      & South \\
      & Artificial lighting       & – \\
      
    \cmidrule(lr){2-3} 
    \multirow[t]{3}{*}{Instrument placement}
      & Non-intrusive camera      & 8MP IMX219, $\times$1 \\
      & Photosensors              & Minolta CL-200A, 4\,$\times$\,4\,$\times$\,3 \\
      & Height of photosensors    & 0.75\,m \\
      
    \cmidrule(lr){2-3} 
    \multirow[t]{2}{*}{Time control}
      & Work schedule             & 8:00\,a.m.\,--\,5:00\,p.m. \\
      & Adjustment intervals      & 5\,min \\
      
    \bottomrule
  \end{tabularx}
\end{table}

\begin{figure}[t]
  \centering
  \includegraphics[width=\linewidth,keepaspectratio]{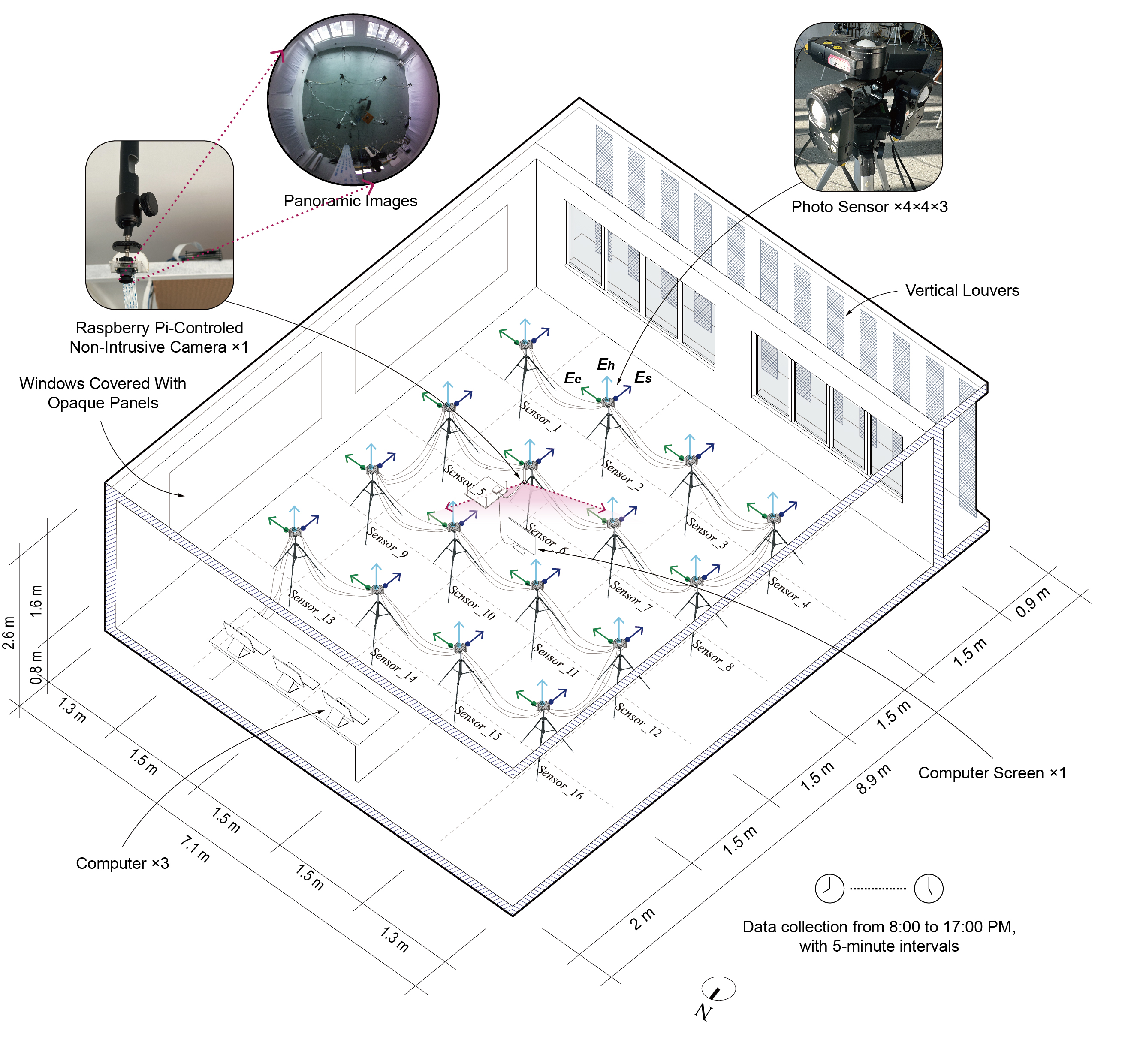}
  \caption{Detailed configuration of the test room.}
  \label{fig:lab}
\end{figure}

As shown in \cref{fig:lab}, the test room is oriented $10^\circ$ east of south, with dimensions of $7.1\,\mathrm{m}\times8.9\,\mathrm{m}\times2.6\,\mathrm{m}$ (L–W–H). There are four identical windows ($2.75\,\mathrm{m}\times1.35\,\mathrm{m}$), two on each of the south and east walls, at a sill height of $0.8\,\mathrm{m}$. To control experimental conditions, the east-facing windows were covered with opaque panels to block all light, while the south-facing windows remained uncovered and closed throughout the test. The south façade is equipped with vertical louvers positioned $0.9\,\mathrm{m}$ from the exterior wall, creating a buffer zone. During the experiment, only natural light from the south-facing windows was used; no artificial lighting was employed.

To capture panoramic images of the side-lit windows from a non-intrusive perspective, a Raspberry Pi–controlled camera (8 MP IMX219) was mounted at the ceiling center. Sixteen measurement points were arranged in a $4\times4$ grid with $1.5\,\mathrm{m}$ spacing. At each point, three photosensors (Minolta CL-200A) on $0.75\,\mathrm{m}$-high tripods measured horizontal illuminance ($E_h$), east-facing vertical illuminance ($E_e$), and south-facing vertical illuminance ($E_s$). Three computers along the north side of the room were each assigned to record data from one illuminance orientation.

\subsubsection{Raw data collection}
\label{subsubsec222}
In this study, data were collected at 5-minute intervals during working hours (from 8:00\,a.m.\ to 5:00\,p.m.) under the laboratory setup. A total of 17,344 data sets were acquired, including both image and tabulated measurements. The image data consist of non-intrusive photographs captured by a ceiling-mounted camera and stored in standard JPEG format. \cref{fig:imagesamples} presents representative images taken at one-hour intervals over a full day, illustrating the variations in side-lit window scenes and indoor daylight conditions.

The constructed data obtained from photosensors were stored in CSV format, comprising timestamps, spatial features, and measured illuminance values ($E_h$, $E_e$, and $E_s$). To enable efficient processing and analysis, both image and constructed data sets were synchronized and labeled according to their timestamps.

\subsection{Deep learning}
\label{subsec23}
\subsubsection{Data-oriented works}
\label{subsubsec231}
The initial dataset used in this study consists of 17,344 sets, including both image and constructed data. To enhance the dataset’s reliability for deep learning, a series of preprocessing steps were applied, including data cleaning, image preprocessing, feature engineering, and dataset partitioning.

\begin{figure}[htbp!]
  \centering

  \includegraphics[width=0.8\linewidth,keepaspectratio]{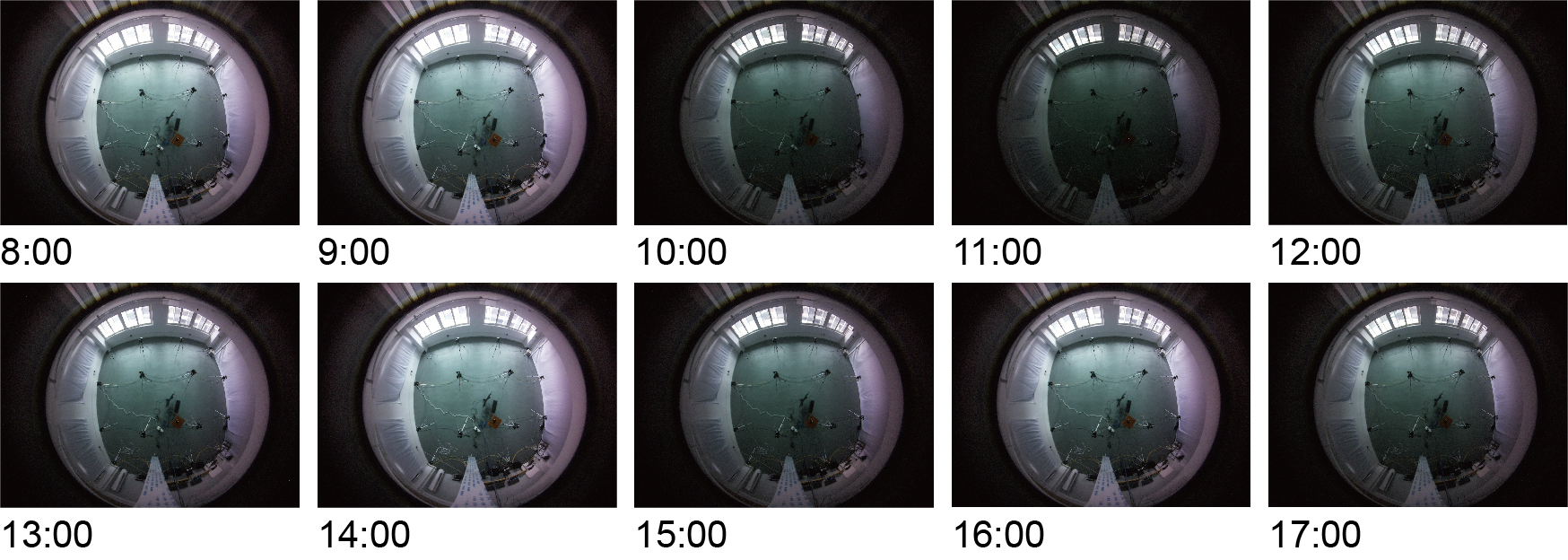}
  \caption{Non-intrusive image samples captured from 8:00 to 17:00.}
  \label{fig:imagesamples}
  
  \vspace{0.5cm} 
  \includegraphics[width=\linewidth,keepaspectratio]{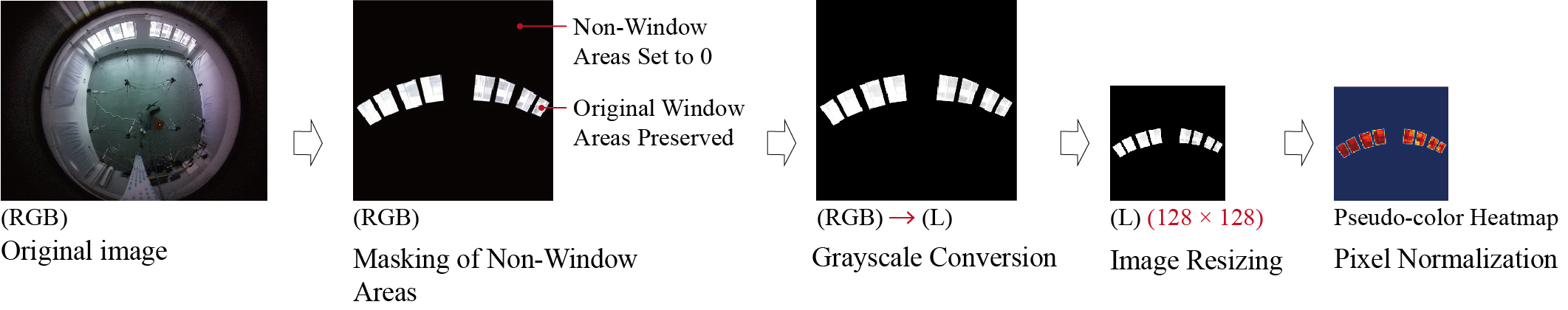}
  \caption{Image pre-processing steps.}
  \label{fig:imageprocessing}
  
\end{figure}

During data collection, occasional interruptions in the photosensor recordings resulted in missing measurements. To ensure data integrity, all samples with missing sensor data were removed. The image pre-processing steps are shown in \cref{fig:imageprocessing}. First, the collected images were masked to zero outside the window area, preserving only the region containing luminance features. Next, all images were converted to grayscale, since intensity in grayscale corresponds to scene luminance. To ensure consistent input dimensions and reduce computational cost, each image was resized to $128\times128$ pixels. Finally, pixel values were normalized to the range $[-1,1]$ to stabilize training and improve convergence.

\begin{figure}[htbp!]
  \centering
  \includegraphics[width=\linewidth,keepaspectratio]{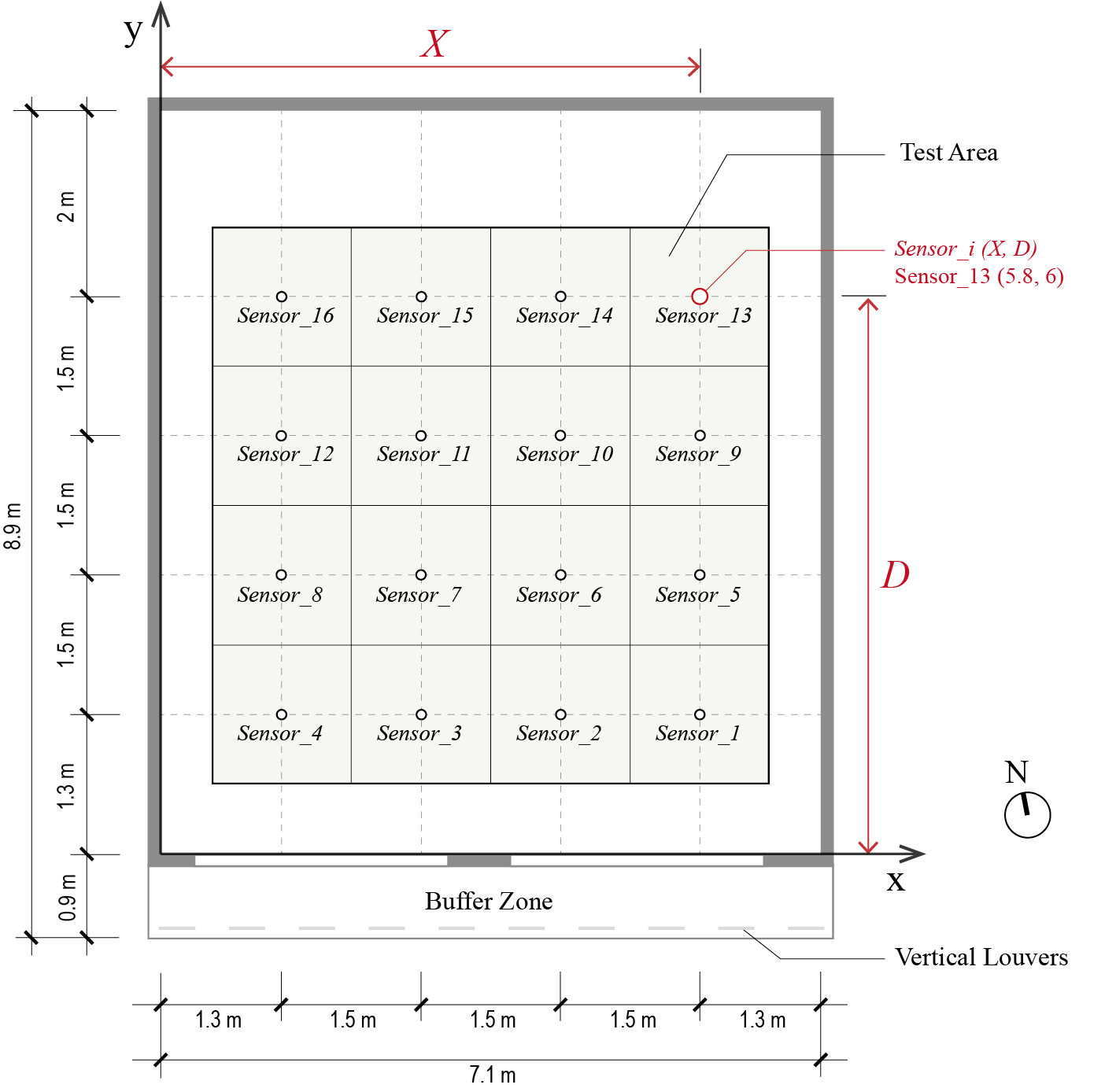}
  \caption{The method for extracting sensor position features.}
  \label{fig:position}
\end{figure}

The feature engineering process focuses on extracting spatial and temporal features. \cref{fig:position} illustrates how spatial features were constructed: a coordinate system was defined where $X$ is the horizontal distance from the sensor to the west wall, and $D$ is the distance from the sensor to the window. Each sensor is labeled as \texttt{Sensor\_i}$(X, D)$. For example, \texttt{Sensor\_13} at $(5.8, 6)$ denotes 5.8 m from the west wall and 6 m from the window. Temporal features were derived from the timestamp by converting each time to minutes since midnight, $t\in[0, 1440)$. To capture the daily cyclical pattern, we then encoded $t$ as two features:

\begin{equation}
  \mathrm{Tod\_sin} = \sin\!\left( \frac{2\pi t}{1440} \right)
  \label{eq:tod_sin}
\end{equation}
\begin{equation}
  \mathrm{Tod\_cos} = \cos\!\left( \frac{2\pi t}{1440} \right)
  \label{eq:tod_cos}
\end{equation}

These two sinusoidal features, $\mathrm{Tod\_sin}$ and $\mathrm{Tod\_cos}$, serve as the encoded temporal inputs to the model.

After data preprocessing, a total of 17{,}344 valid data samples were obtained. To assess the model’s generalization to unseen temporal scenarios, all data collected on the final day (1{,}744 samples) were assigned to an independent testing set (Test Set 2). The remaining 15{,}600 samples were randomly divided using stratified sampling: 70\% were assigned to the training set (10{,}920 samples), and the remaining 30\% were equally split into a validation set (2{,}340 samples) and a testing set (Test Set 1, 2{,}340 samples). A fixed random seed was applied to ensure reproducibility.

In addition, z-score normalization was conducted on both input features and output labels to standardize them to zero mean and unit variance. The scaling parameters (mean and standard deviation) were fitted on the training set and then applied to the validation and testing sets to prevent data leakage.

\subsubsection{Model-oriented works}
\label{subsubsec232}
The objective of this study is to develop a non-intrusive, image-based method for real-time prediction of the indoor daylight environment. The task is formulated as a multi-output regression problem that combines two input modalities: image features and structured features (temporal and spatial), to predict three directional illuminance values ($E_h$, $E_s$, and $E_e$) at a specified point in space in real time. CNNs are highly effective at processing image information ~\cite{o2015introduction} and have been widely applied to various spatial learning tasks via stacked convolutional layers ~\cite{li2021survey}. MLPs are classical neural network architectures designed to model complex non-linear relationships among variables. Previous studies have demonstrated that these algorithms can achieve strong performance in building performance prediction ~\cite{he2021predictive,yan2021graph}.

\begin{figure}[htbp!]
  \centering
  \includegraphics[width=\linewidth,keepaspectratio]{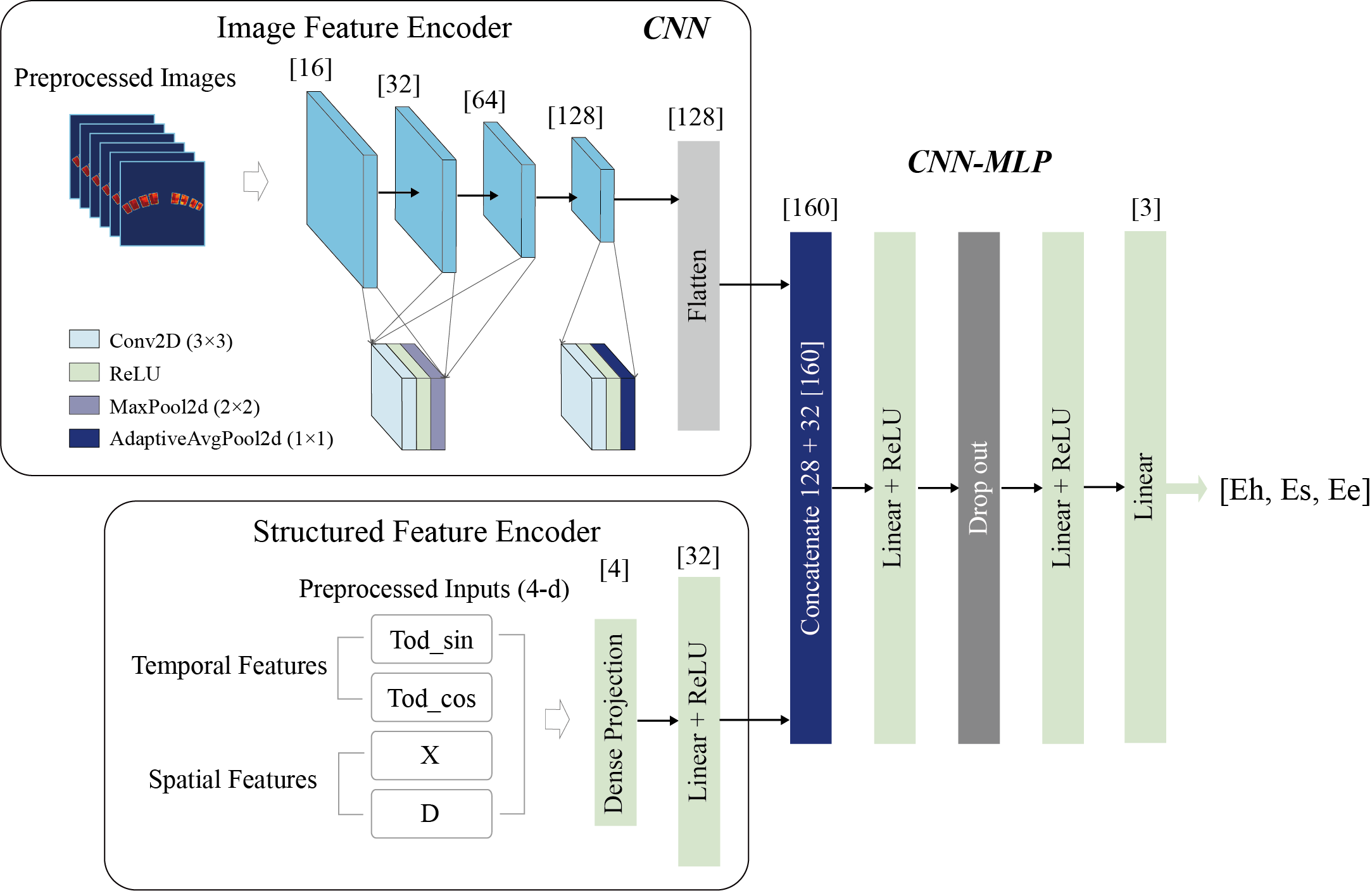}
  \caption{Architecture of the proposed multimodal CNN–MLP model.}
  \label{fig:multimodal_model}
\end{figure}

To effectively integrate both image and structured inputs, we adopt a multimodal CNN–MLP architecture as shown in \cref{fig:multimodal_model}. The model takes two types of inputs including preprocessed images of side-lit windows and four-dimensional structured features. The images are encoded into 128-dimensional feature vectors by a four-layer CNN with channel sizes [16, 32, 64, 128]. The structured features are projected into a 32-dimensional embedding. These two embeddings are then concatenated into a 160-dimensional fused representation, which is fed into an MLP regressor. The model outputs three directional illuminance values ($E_h$, $E_s$, and $E_e$).

To evaluate the performance of the model, the dataset was divided into training, validation, and two distinct testing sets (Test Set 1 and Test Set 2) as described previously. The validation set was used to monitor model performance during training and to guide hyperparameter tuning. Test Set 1 was used for preliminary evaluation on data drawn from the same distribution as the training set, while Test Set 2, containing only samples from the final day, was used to assess generalization to temporally unseen data.

Hyperparameters were manually tuned based on validation performance and empirical observations of training stability. This study focused on two key hyperparameters: the MLP architecture (i.e., number of layers and hidden units) and the dropout rate applied to the MLP block. Other hyperparameters including learning rate, batch size, structured feature embedding dimensionality, and the configuration of the CNN encoder were held constant across experiments to ensure fair comparisons.

Model performance was assessed using three standard regression metrics: coefficient of determination ($R^2$), mean absolute error (MAE), and root mean squared error (RMSE)~\cite{goodfellow2016deep}. $R^2$ quantifies the proportion of variance explained by the model, MAE measures the average absolute prediction error (robust to outliers), and RMSE penalizes larger errors more heavily.

To ensure physical interpretability, both predicted and ground truth values were inverse-transformed using the fitted scaler before evaluation. All performance metrics were computed in the original physical unit of illuminance (lux). Model evaluation was conducted on two separate testing sets: Test Set 1, drawn from the same distribution as the training data, and Test Set , which comprised temporally distinct samples collected on the final day. This design allowed for a comprehensive assessment of both within-distribution accuracy and generalization to temporally unseen data.

\section{Results}
\label{sec3}
\subsection{Dataset overview and feature analysis}
\label{subsec31}
\begin{figure}[htbp!]
  \centering
  \includegraphics[width=0.7\linewidth,keepaspectratio]{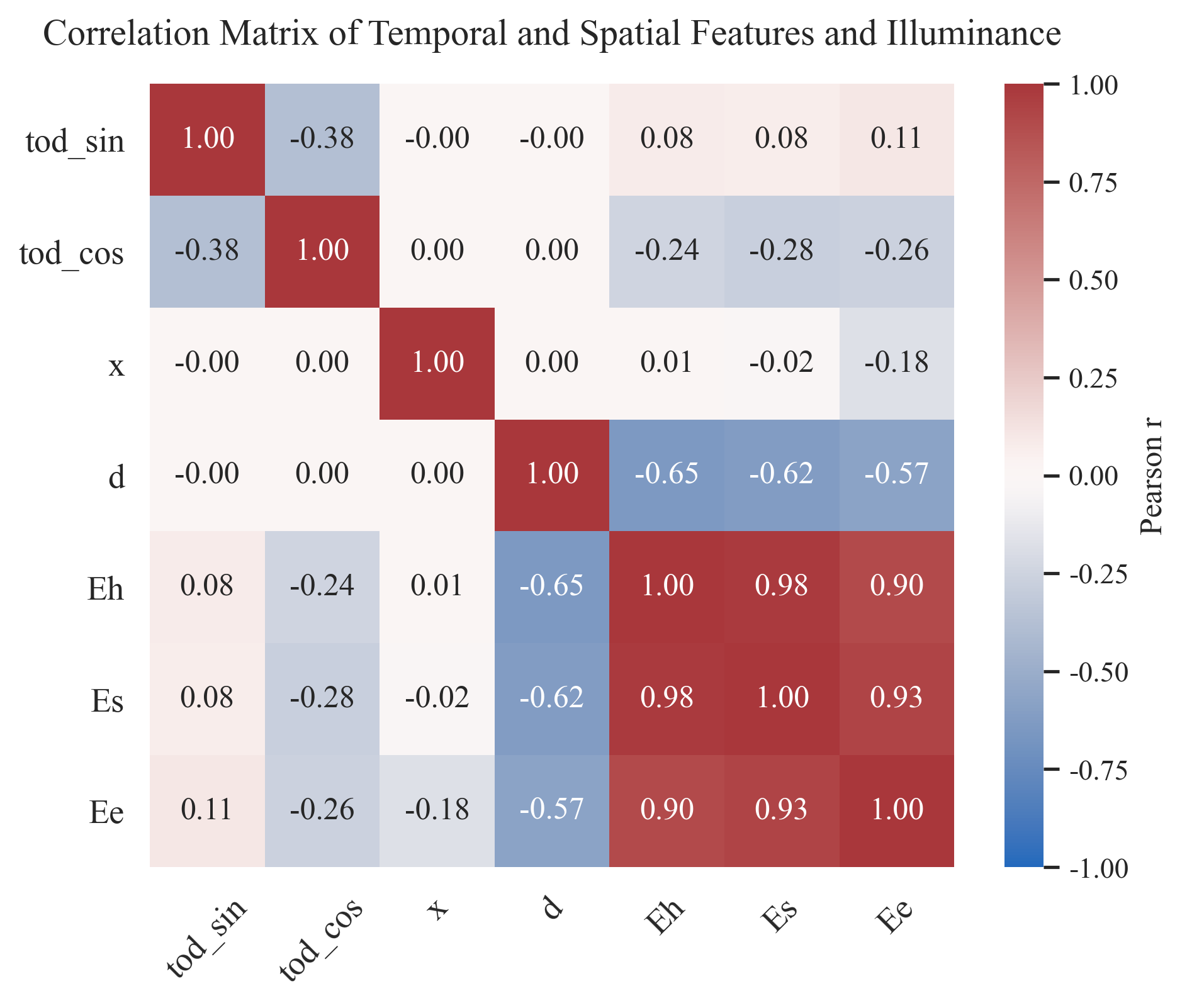}
  \caption{Correlation matrix of temporal and spatial features and illuminance.}
  \label{fig:correlation}
\end{figure}

In this study, the correlation matrix was used to analysis the initial validation of the collected dataset and guide the model design. ~\cref{fig:correlation} presents the Pearson correlation matrix between the engineered temporal features (\(\mathrm{Tod\_sin}\), \(\mathrm{Tod\_cos}\)) and spatial features (\(X\), \(D\)) and the three output illuminance variables (\(E_h\), \(E_s\), \(E_e\)).

The three output illuminance variables exhibit strong positive correlations with one another (\(r>0.9\)). This indicates strong multicollinearity and redundancy among the outputs, suggesting that they share numerous underlying patterns. Accordingly, a multi-output regression model would be more suitable, as it can capture the shared patterns between the targets through a common feature extraction process.

In terms of temporal features, \(\mathrm{Tod\_cos}\) exhibits moderate negative correlations with the illuminance variables (approximately \(r = -0.24\) to \(-0.28\)). This pattern reflects the unimodal diurnal cycle of daylight, with peak illuminance occurring at solar noon, corresponding to the minimum value in the cosine-based temporal encoding. In contrast, \(\mathrm{Tod\_sin}\) shows weak correlations with all target variables (\(r<0.12\)). This may be attributed to the relatively symmetric distribution of daylight between morning and afternoon during the data collection period, leading to opposing effects that offset each other in a global linear analysis.

For the spatial features, \(D\) shows strong negative correlations with all illuminance variables (\(r = -0.57\) to \(-0.65\)), consistent with physical expectations: as the distance from the window increases, the amount of incident daylight decreases significantly. On the other hand, \(X\) exhibits very weak linear correlations (ranging from \(-0.18\) to \(0.01\)), suggesting limited linear influence. However, it is worth noting that such weak correlations do not preclude the existence of nonlinear effects or feature interactions, particularly under specific spatial layouts or temporal conditions, which may still be captured by the deep learning model.

\begin{figure}[t]
  \centering
  \includegraphics[width=0.8\linewidth,keepaspectratio]{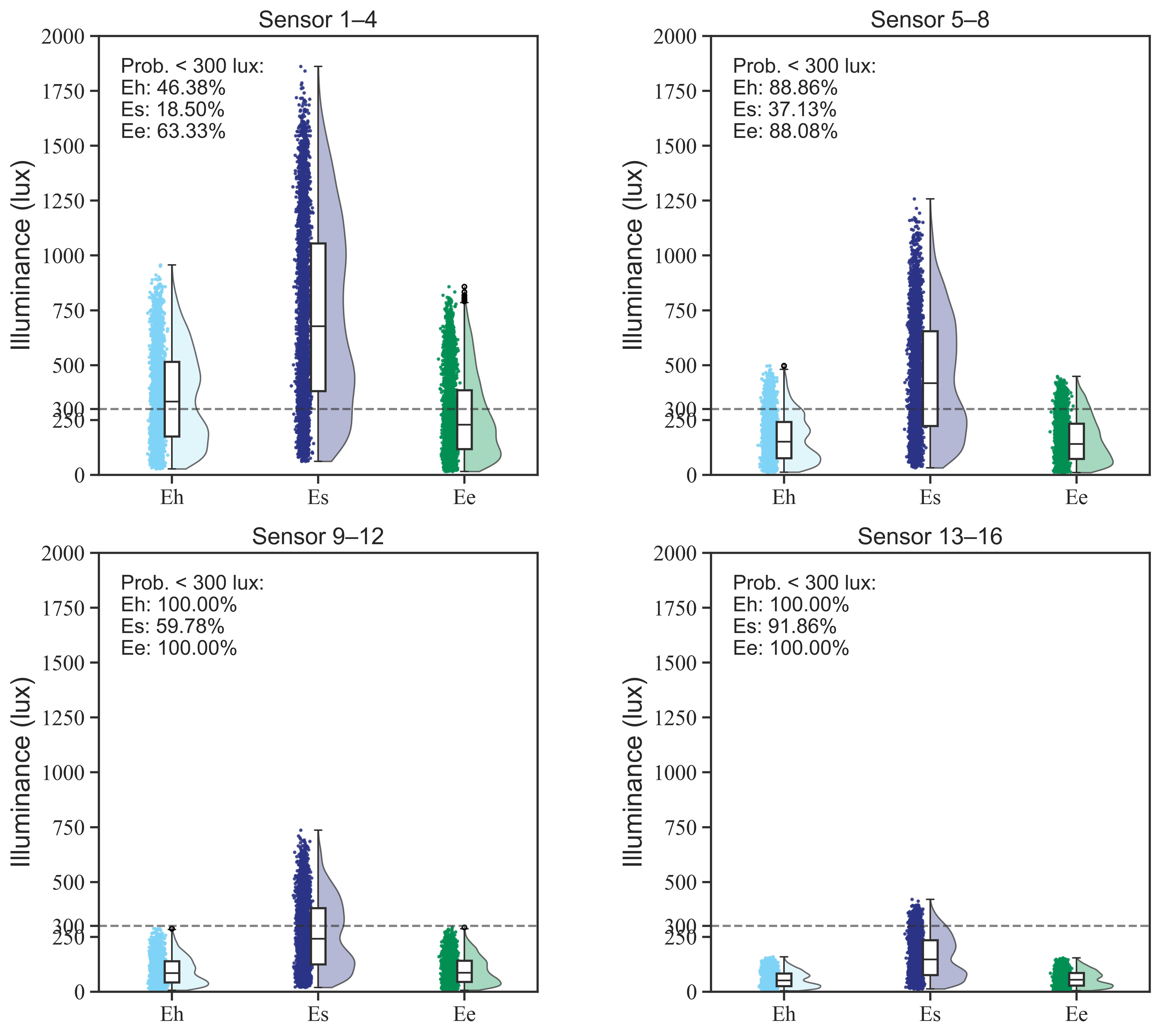}
  \caption{Spatial distribution of illuminance indicators (Eh, Es, Ee) across sensor zones.}
  \label{fig:illuminance_dist}
\end{figure}

~\cref{fig:illuminance_dist} illustrates the illuminance distributions of 16 sensor points in three directions (\(E_h\), \(E_s\), \(E_e\)). Among these, \(E_s\) consistently exhibits higher illuminance levels and a wider distribution range compared to \(E_h\) and \(E_e\), with more pronounced fluctuations. This trend might be attributed to the south-facing orientation, which generally leads to greater direct solar exposure. A clear spatial gradient is observed from the window-adjacent area (Sensors 1–4) to the deeper zone (Sensors 13–16), with overall illuminance levels gradually decreasing. Furthermore, both \(E_h\) and \(E_e\) exhibit narrow distributions centered around lower illuminance levels, with values at Sensors 9–16 in both directions almost entirely below 300 lux. This is likely because both are mainly influenced by diffuse daylight.

On the other hand, it is worth noting that the illuminance distributions in all three directions exhibit a certain degree of skewness and a few outliers. \(E_h\) and \(E_e\) values are primarily distributed in the low illuminance range (\(<300\) lux), while high-illuminance samples (\(>1000\) lux) are extremely rare. Although \(E_s\) shows higher illuminance levels, its distribution remains skewed toward the mid-to-low range. In addition, outliers are observed across all directions, mainly distributed at high illuminance levels. This pattern is especially prominent for \(E_s\), with some values reaching as high as 1500–2000 lux. These limitations may lead to reduced predictive accuracy under high illuminance levels during model training. To improve the performance and generalization ability of the model, several techniques should be adopted, including sampling balance strategies, robust loss functions, and regularization.

\begin{figure}[htbp!]
  \centering
  \includegraphics[width=\linewidth,keepaspectratio]{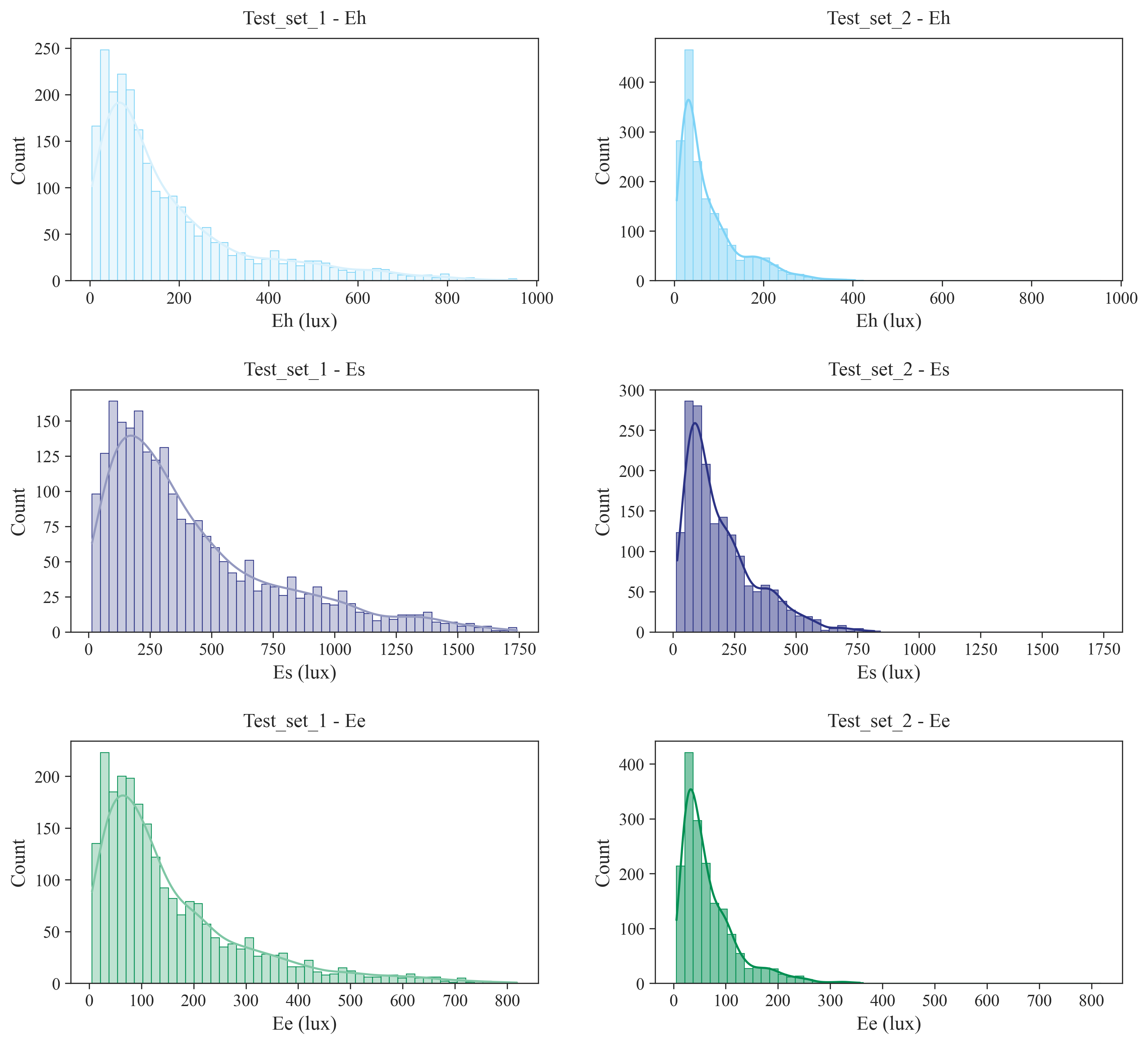}
  \caption{Data distribution of illuminance indicators (Eh, Es, Ee) on Test Set 1 and Test Set 2.}
  \label{fig:test_illuminance_dist}
\end{figure}

As mentioned in the previous section, Test Set~1 shares the same data distribution as the training set through stratified sampling, while Test Set~2 consists of data collected on a new day. ~\cref{fig:test_illuminance_dist} presents the distribution of three illuminance indicators (\(E_h\), \(E_s\), \(E_e\)) across the two test sets. Although the illuminance values in Test Set~2 still fall entirely within the range covered by Test Set~1, the values are concentrated in the low-to-mid illuminance range. High-illuminance samples such as those exceeding 1000\,lux are absent in Test Set~2. This domain shift may provide an explanation for the observed degradation in model performance on Test Set~2.

\subsection{Model selection and hyperparameter tuning}
\label{subsec32}
Among the various hyperparameters involved in model training, this study focuses on tuning the architecture of the MLP module and the dropout rate applied to it. These two factors were selected because they directly impact the model’s capacity and regularization, which are critical for balancing prediction accuracy and generalization ability. The MLP module serves to integrate image and structured features and produce the final illuminance prediction. The number of layers and hidden nodes are key architectural factors that determine the model’s capacity and expressiveness. Dropout is a widely adopted regularization technique that mitigates overfitting by randomly deactivating neurons during training. This effect is particularly important in deeper MLP architectures, where overfitting is more likely to occur. Other hyperparameters were fixed during all experiments as listed in ~\Cref{tab:hyperparams}. These settings were determined through preliminary experiments and were shown to provide stable training performance.

\begin{table}[htbp!]
  \centering
  
  \renewcommand{\baselinestretch}{1.0}\selectfont  
  \scriptsize                                      
  \renewcommand{\arraystretch}{1.2}               
  \setlength{\tabcolsep}{6pt}                     
  
  \caption{The setting of hyperparameters.}
  \label{tab:hyperparams}
  
  \begin{tabular}{ll}
    \toprule  
    Hyperparameter & Setting \\ 
    \midrule
    Learning rate                       & 0.001 \\
    Batch size                          & 64 \\
    Structured feature projection size  & 32 \\
    CNN channels per layer              & [16, 32, 64, 128] \\
    Optimizer                           & Adam \\
    Early stopping patience             & 12 \\
    Maximum epochs                      & 200 \\
    \bottomrule 
  \end{tabular}
  
\end{table}

\begin{figure}[htbp!]
  \centering
  \begin{subfigure}[b]{0.48\textwidth}
    \includegraphics[width=\textwidth]{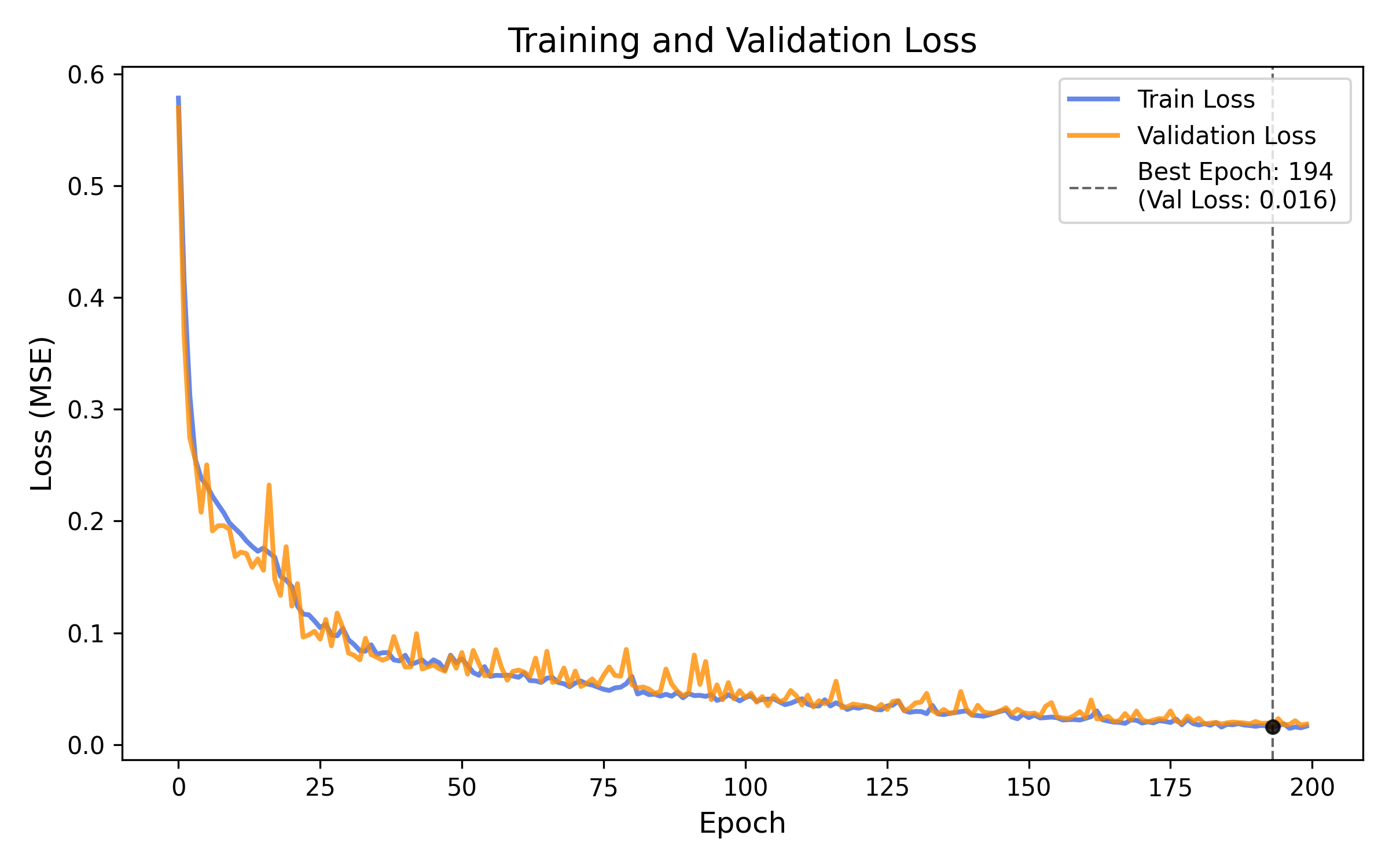}
  \end{subfigure}
  \hfill
  \begin{subfigure}[b]{0.48\textwidth}
    \includegraphics[width=\textwidth]{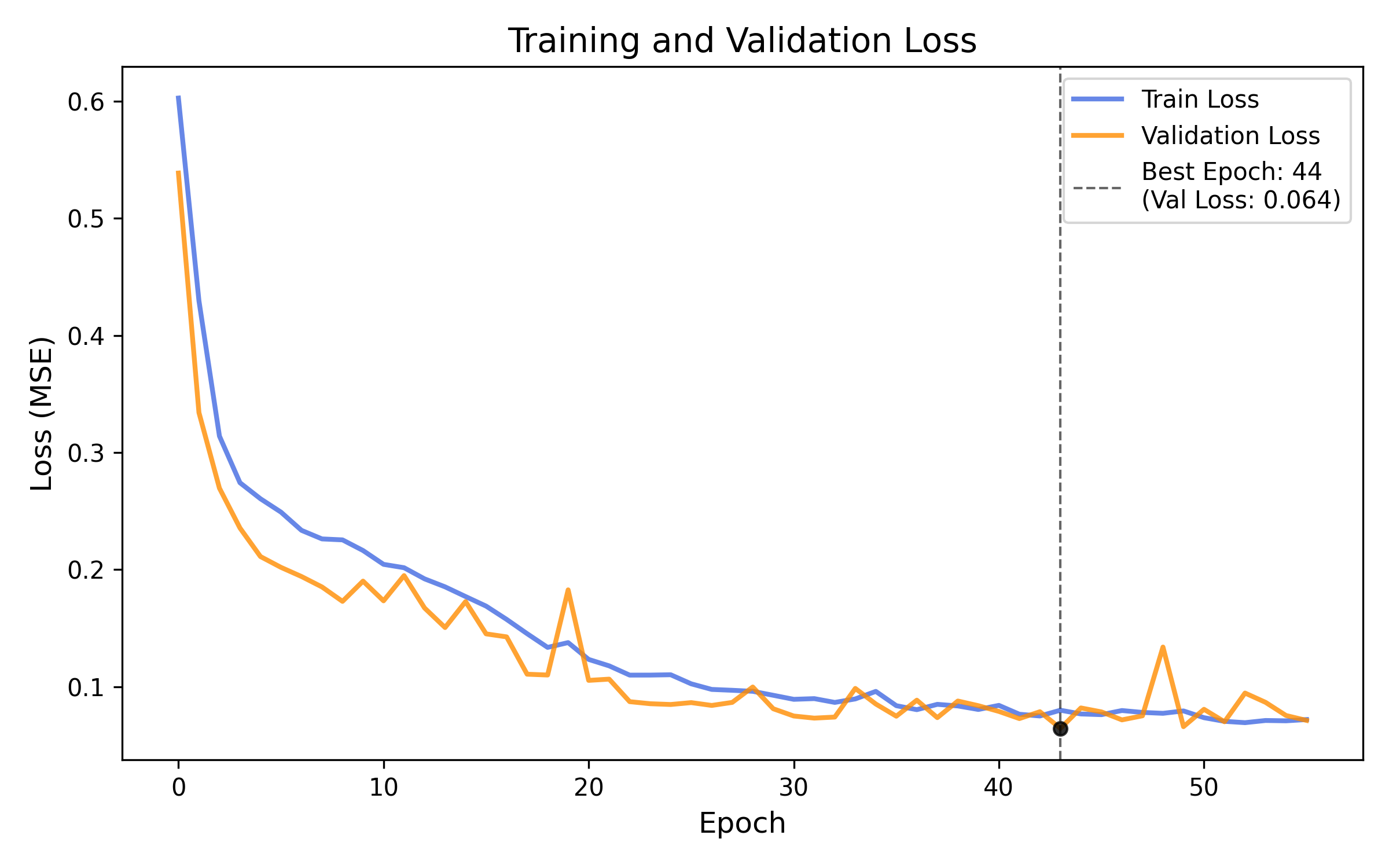}
  \end{subfigure}

  \begin{subfigure}[b]{0.48\textwidth}
    \includegraphics[width=\textwidth]{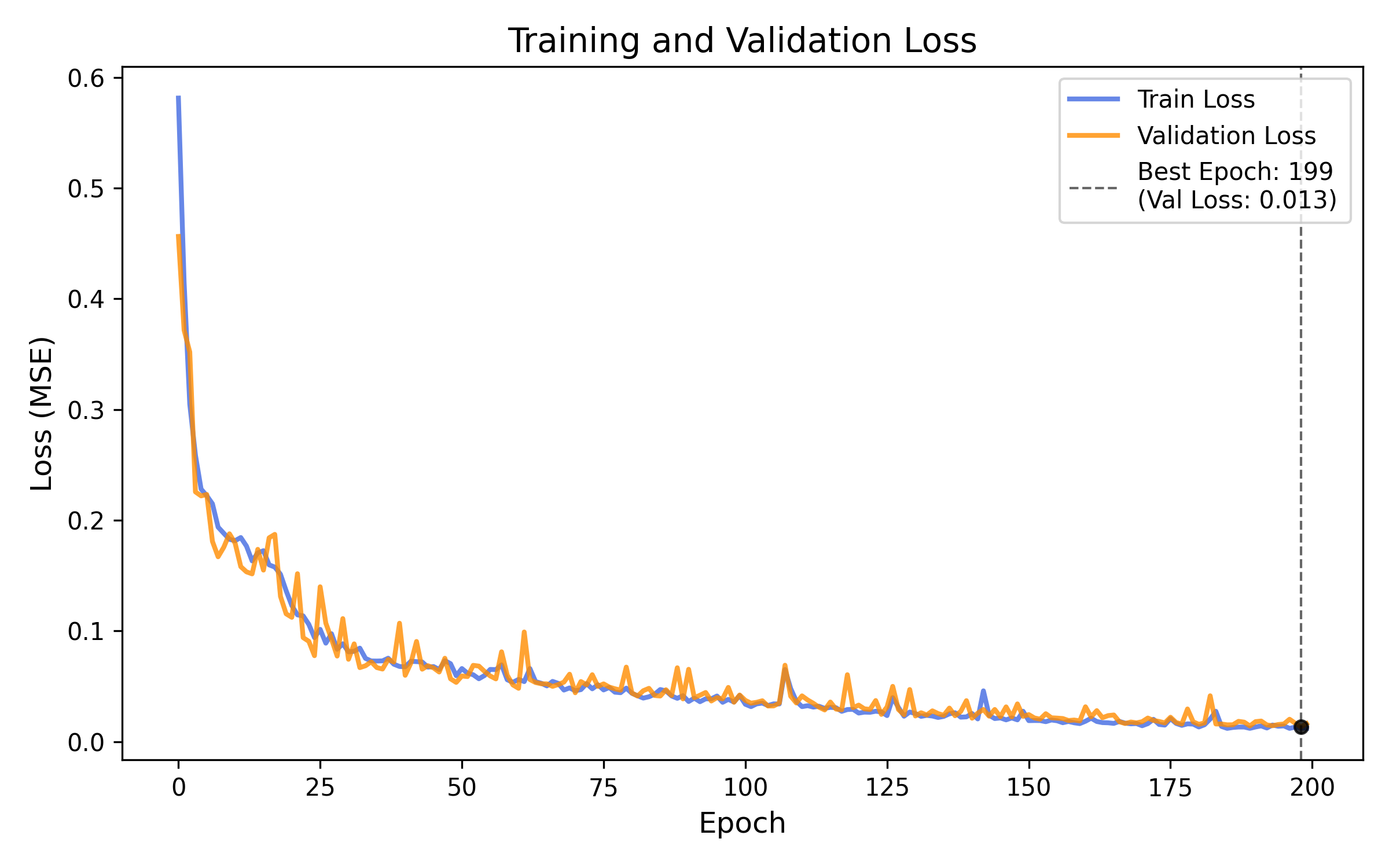}
  \end{subfigure}
  \hfill
  \begin{subfigure}[b]{0.48\textwidth}
    \includegraphics[width=\textwidth]{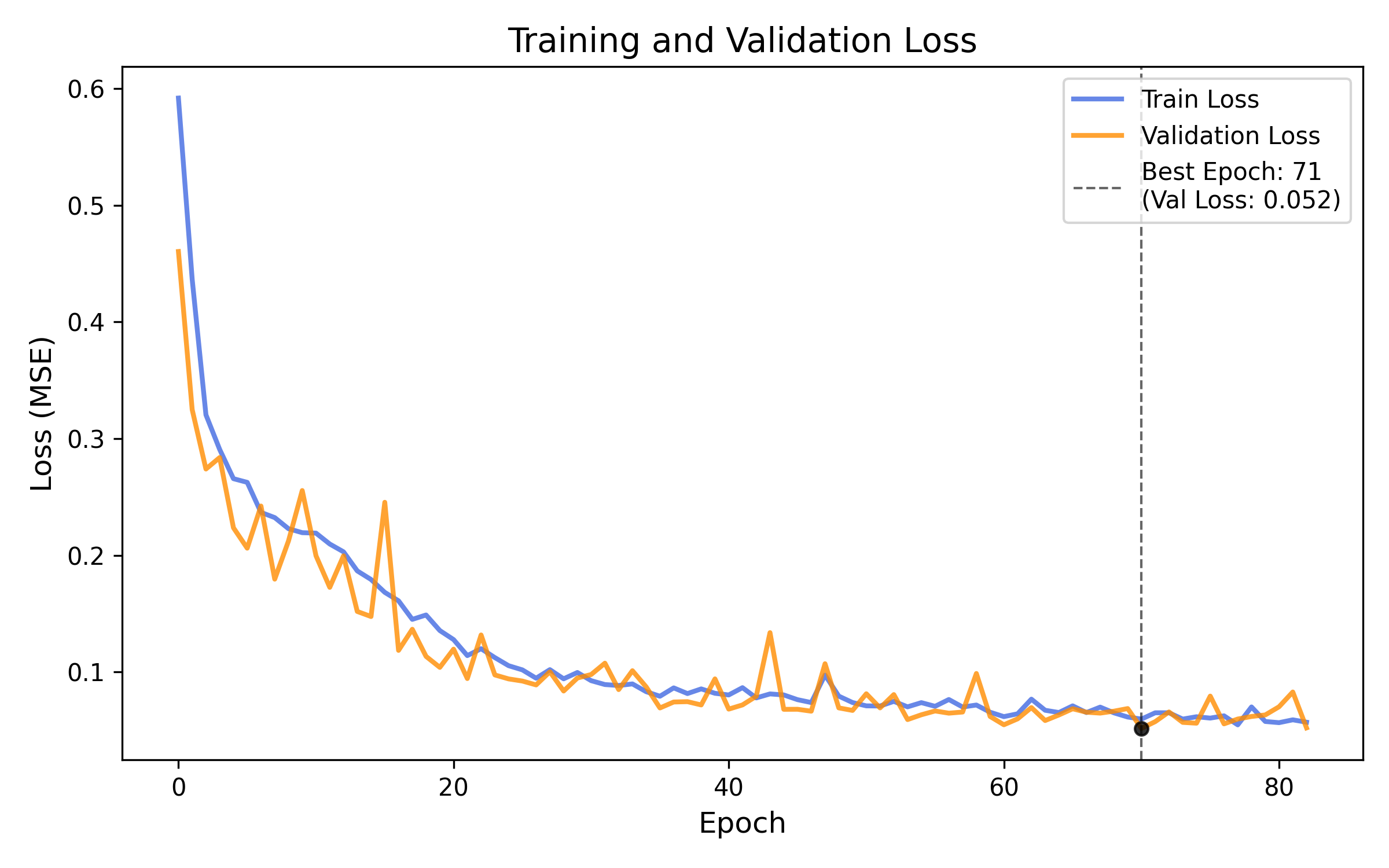}
  \end{subfigure}

  \begin{subfigure}[b]{0.48\textwidth}
    \includegraphics[width=\textwidth]{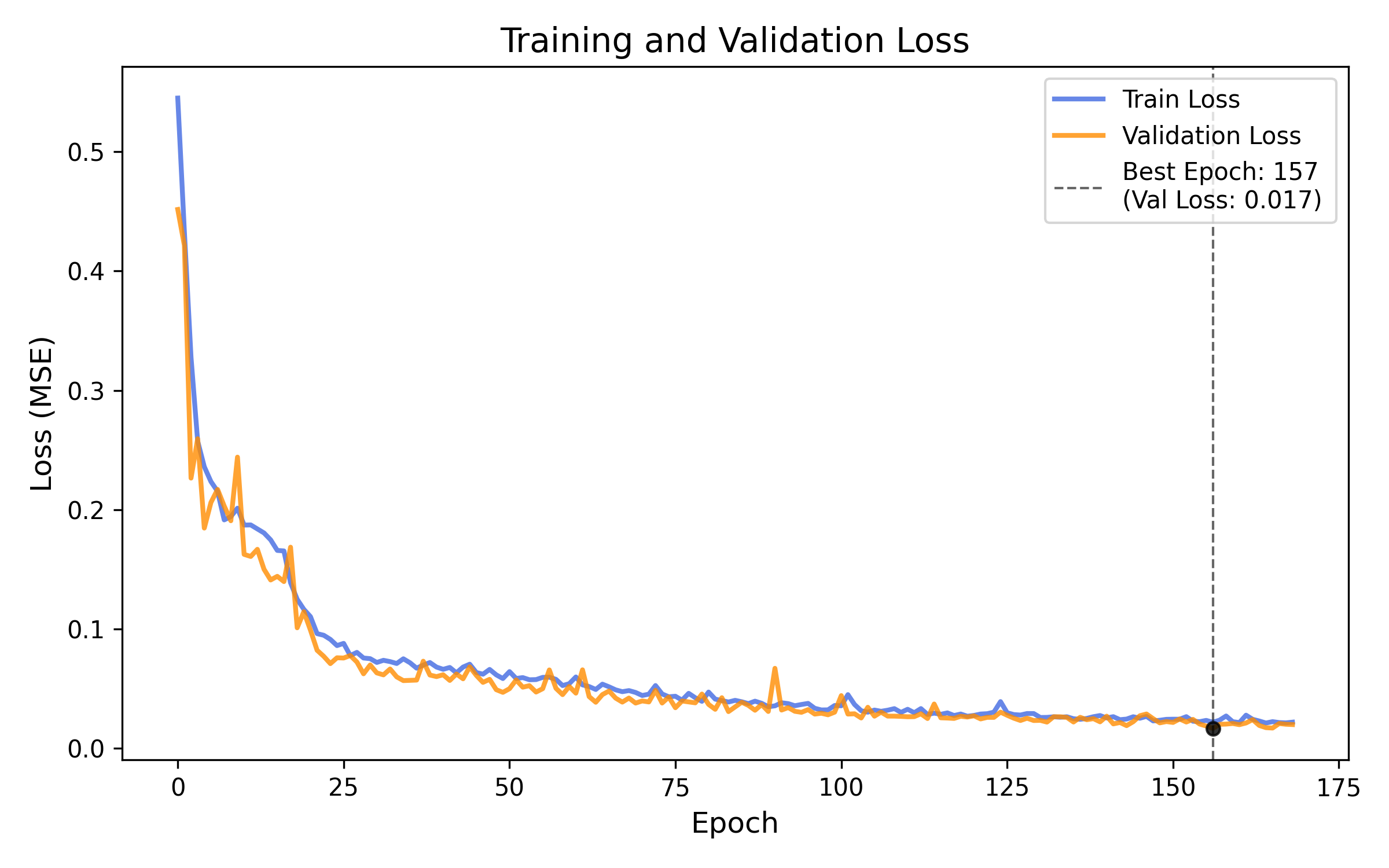}
  \end{subfigure}
  \hfill
  \begin{subfigure}[b]{0.48\textwidth}
    \includegraphics[width=\textwidth]{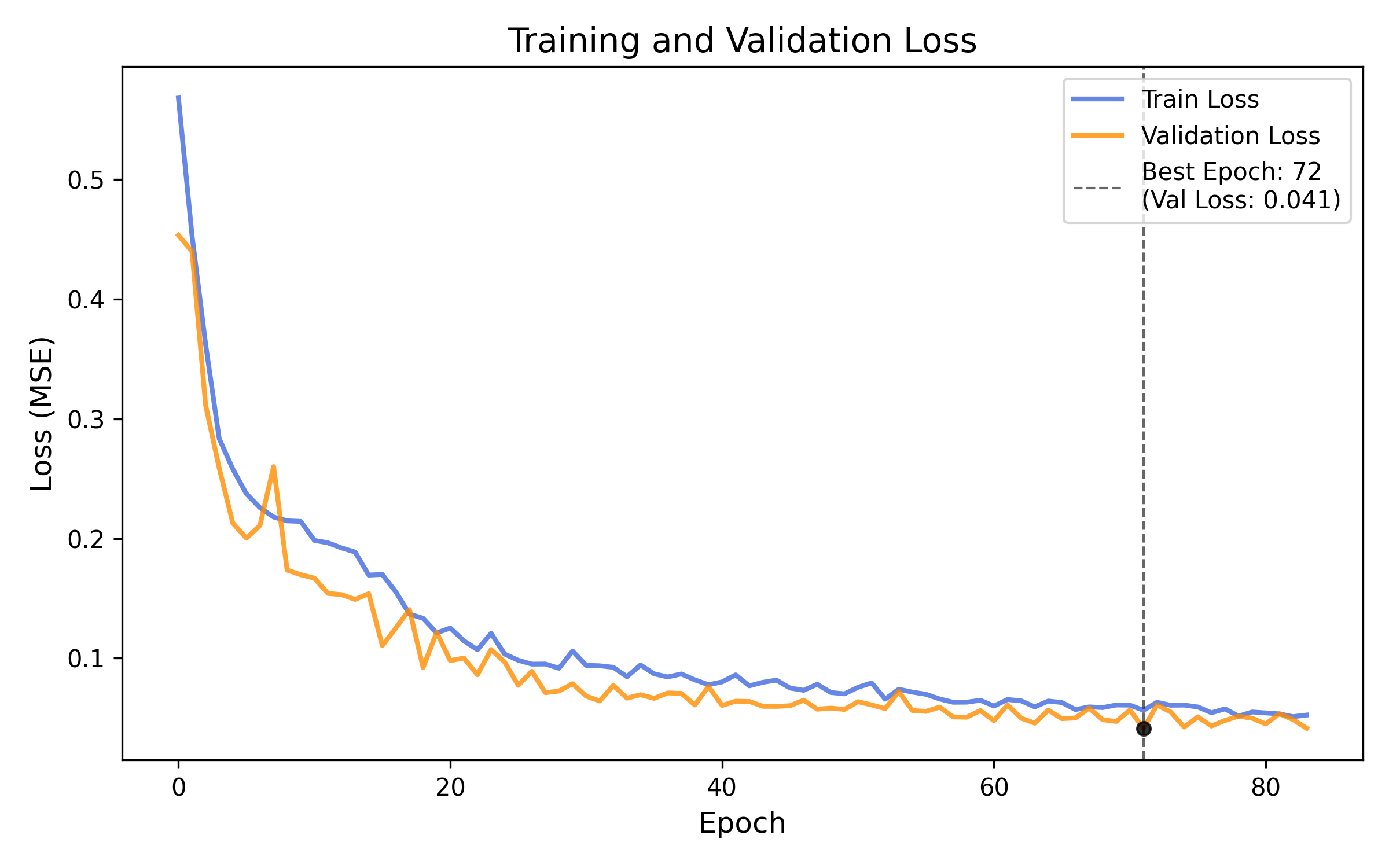}
  \end{subfigure}

  \caption{Learning curves of the six candidate models on the training and validation sets.}
  \label{fig:learning_curves}
\end{figure}

Based on this, six candidate configurations with varying MLP structure and dropout rate were tested during hyperparameter tuning. ~\Cref{tab:configurations} summarizes the details of each configuration along with their validation performance, including mean squared error (MSE), coefficient of determination ($R^2$), and the best epoch determined by early stopping. The MLP structures range from shallow architectures with no regularization (\([64,32]\), dropout = 0.0) to deeper networks with stronger dropout (\([256,128,64]\), dropout = 0.5). In addition, ~\cref{fig:learning_curves} further illustrates the training and validation loss curves for each configuration.

Among the six candidates, Model C achieved the best validation performance, with the lowest MSE of 0.013 and the highest $R^2$ of 0.987. Both the training and validation loss curves exhibited a consistent and steadily decreasing pattern without noticeable fluctuations, with the optimal validation performance reached at the final epoch (epoch 199). Model A and Model E also demonstrated comparable results, with validation MSEs of 0.016 and 0.017 and $R^2$ values of 0.984 and 0.983, respectively. These models achieved validation performance within a narrow range, with $R^2$ consistently above 0.98 and MSE below 0.018, indicating strong predictive accuracy during training. This suggests that the model benefits from a moderately deep structure with sufficient representational capacity, while not requiring additional regularization to prevent overfitting.

Moreover, despite applying the same early stopping criterion (patience = 12) across all models, those with higher dropout rates, such as Model B, Model D, and Model F, tended to reach this threshold significantly earlier than their non-regularized counterparts. In particular, Model B and Model D not only converged earlier but also introduced marked oscillations in the validation loss curve, reflecting unstable learning dynamics.

While several candidate models, including Model A, Model C, and Model E, achieved comparable validation performance, Model C offered the most favorable balance among predictive accuracy, training stability, and architectural complexity. Its moderately deep structure, without the use of dropout regularization, provides greater representational capacity than shallower alternatives while avoiding the instability and potential over-regularization observed in deeper configurations. Based on these considerations, Model C was selected as the final model for subsequent evaluation and comparative analysis.

\begin{table}[htbp!]
  \centering
  \caption{MSE and $R^2$ of six candidate models during hyperparameter tuning.}
  \label{tab:configurations}

  \begin{tabular}{l l l l l l l}
    \toprule
    Model & MLP Structure & Dropout & Learning rate & Val MSE & Val $R^2$ & Best epoch \\
    \midrule
    Model A & [64, 32]      & 0.0 & 0.001 & 0.016 & 0.984 & 194 \\
    Model B & [64, 32]      & 0.3 & 0.001 & 0.065 & 0.935 &  44 \\
    Model C & [128, 64, 32] & 0.0 & 0.001 & 0.013 & 0.987 & 199 \\
    Model D & [128, 64, 32] & 0.3 & 0.001 & 0.052 & 0.948 &  71 \\
    Model E & [256, 128, 64]& 0.3 & 0.001 & 0.017 & 0.983 & 157 \\
    Model F & [256, 128, 64]& 0.5 & 0.001 & 0.041 & 0.959 &  72 \\
    \bottomrule
  \end{tabular}
  
\end{table}

\subsection{Test performance on unseen data}
\label{subsec33}
\begin{figure}[htbp!]
  \centering
  \includegraphics[width=\linewidth,keepaspectratio]{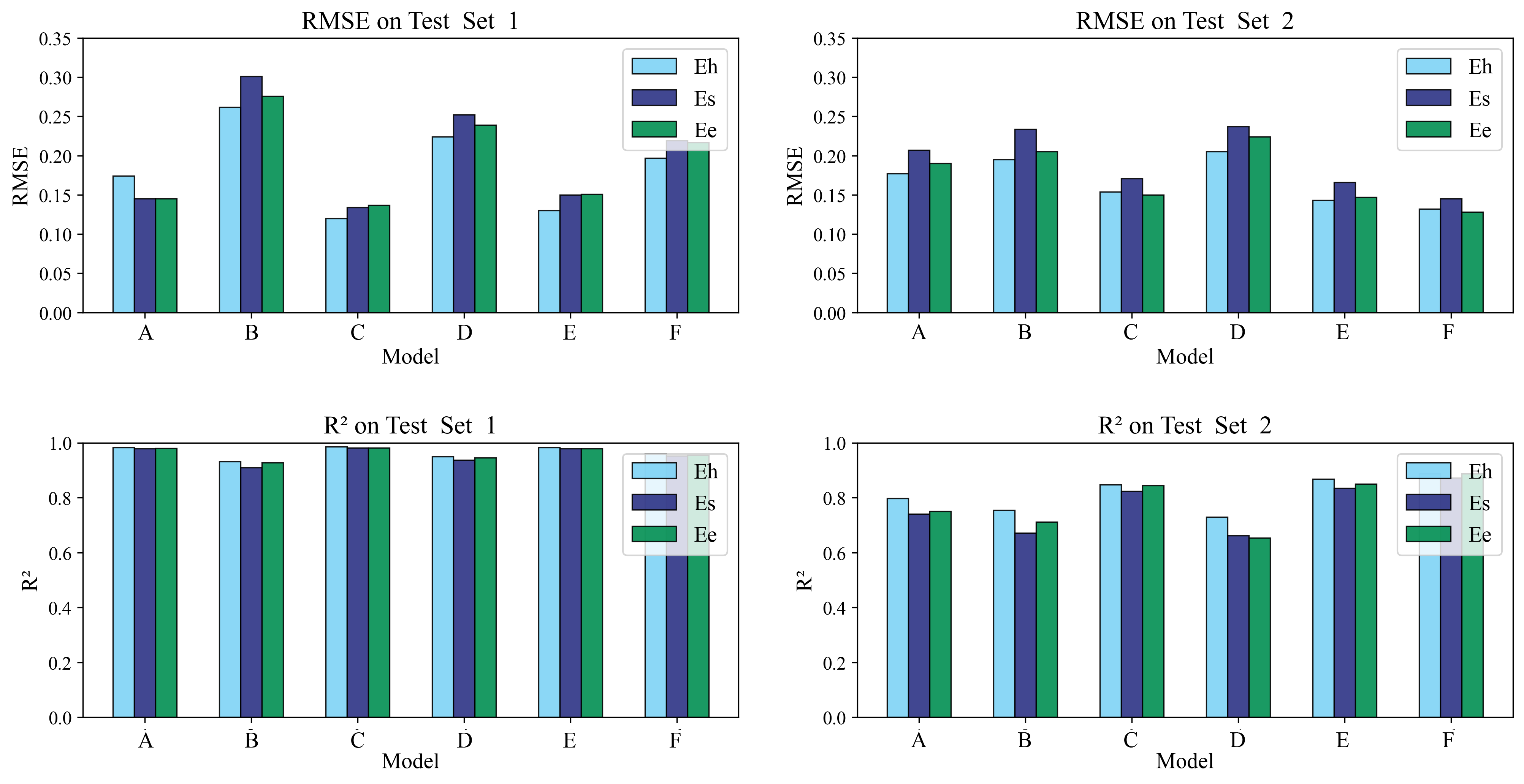}
  \caption{Comparison of RMSE and $R^2$ values across six models on Test Set 1 and Test Set 2.}
  \label{fig:test_performance}
\end{figure}

To further evaluate the generalization ability of the selected model (Model C), its performance was assessed on two independent test sets, with a comparative evaluation made against the remaining configurations. ~\Cref{tab:test_performance} and ~\cref{fig:test_performance} summarize the test performance of all six models on the two test sets using RMSE and $R^2$ metrics for the prediction of $E_h$, $E_s$, and $E_e$. Overall, most models performed better on Test Set 1, with RMSE values ranging from 0.120 to 0.301 and $R^2$ values between 0.931 and 0.986, reflecting strong predictive accuracy. In contrast, Test Set 2 presented more pronounced generalization challenges, as RMSE values ranged from 0.132 to 0.237, and $R^2$ scores dropped to as low as 0.653 in certain models. While performance declined on Test Set 2 across all models, some models retained reasonably good predictive accuracy. Moreover, the predictive performance for the three target variables was generally comparable, with a slight decrease observed for $E_s$.

Among the six candidates, Model C and Model E achieved the best overall performance across both test sets, consistently yielding the lowest RMSE values and the highest $R^2$ scores for all three target variables. On Test Set 1, both models achieved $R^2$ values above 0.978 for all outputs. Moreover, their performance remained robust on Test Set 2, with RMSE values consistently below 0.17 and $R^2$ values above 0.82, indicating acceptable generalization ability. It is also worth noting that Model F, although less competitive on Test Set 1, performed well on Test Set 2, achieving the lowest RMSE and the highest $R^2$ for each target variable among all models.

In contrast, Model A, Model B, and Model D demonstrated a noticeable decline in performance on Test Set 2, with $R^2$ scores falling below 0.75 across most targets, indicating reduced generalization capacity. The combination of limited depth and aggressive dropout appears to have impaired these models’ ability to generalize to unseen data domains. The shallow architecture of Model A likely lacked sufficient representational capacity, limiting its ability to capture the underlying patterns in more varied or complex data. Meanwhile, Model B and Model D incorporated dropout regularization, which was originally intended to prevent overfitting but may have disrupted learning by inducing early convergence and reducing the model’s effective capacity. As shown in ~\Cref{tab:configurations}, these two models reached early stopping after only 44 and 71 epochs, respectively, suggesting that training may have been terminated before sufficient optimization was achieved. This insufficient training could have further contributed to their weakened generalization on Test Set 2.

Furthermore, several models, such as Model B, Model D, and Model F, exhibited lower RMSE values on Test Set 2 compared to Test Set 1. For instance, Model F achieved a reduction in RMSE from 0.197 to 0.132 for $E_h$ prediction, and similar reductions were observed for $E_s$ and $E_e$. Although the model exhibited lower RMSE values on Test Set 2, their $R^2$ values also declined. This may be attributed to the narrower distribution and reduced variance of target values in Test Set 2 as shown in ~\cref{fig:test_illuminance_dist}, which likely led to smaller absolute errors.

\newcolumntype{L}[1]{>{\raggedright\arraybackslash}m{#1}}

\begin{table}[htbp!]
  \centering
  \renewcommand{\baselinestretch}{1.0}\selectfont
  \scriptsize 
  
  \renewcommand{\arraystretch}{1.1}
  \setlength{\tabcolsep}{4pt} 
  
  \caption{RMSE and $R^2$ values of $E_h$, $E_s$, and $E_e$ prediction for six models on Test Set 1 and Test Set 2.}
  \label{tab:test_performance}
  
  \begin{tabular}{@{}L{1.6cm} L{1.8cm} *{3}{L{1.2cm}} *{3}{L{1.2cm}}@{}}
    \toprule
    
    Model & Test Set 
      & \multicolumn{3}{l}{RMSE} 
      & \multicolumn{3}{l}{$R^2$} \\
    
    \cmidrule(lr){3-5} \cmidrule(lr){6-8}
    & 
      & $E_h$ & $E_s$ & $E_e$ 
      & $E_h$ & $E_s$ & $E_e$ \\
    \midrule
    \multirow[t]{2}{*}{Model A}
      & Test Set~1 & 0.174 & 0.145 & 0.145 & 0.983 & 0.979 & 0.980 \\
    \cmidrule(l){2-8}
      & Test Set~2 & 0.177 & 0.207 & 0.190 & 0.797 & 0.741 & 0.750 \\
    \midrule
    \multirow[t]{2}{*}{Model B}
      & Test Set~1 & 0.262 & 0.301 & 0.276 & 0.931 & 0.910 & 0.927 \\
    \cmidrule(l){2-8}
      & Test Set~2 & 0.195 & 0.234 & 0.205 & 0.754 & 0.671 & 0.711 \\
    \midrule
    \multirow[t]{2}{*}{Model C}
      & Test Set~1 & 0.120 & 0.134 & 0.137 & 0.986 & 0.982 & 0.982 \\
    \cmidrule(l){2-8}
      & Test Set~2 & 0.154 & 0.171 & 0.150 & 0.847 & 0.824 & 0.844 \\
    \midrule
    \multirow[t]{2}{*}{Model D}
      & Test Set~1 & 0.224 & 0.252 & 0.239 & 0.950 & 0.937 & 0.945 \\
    \cmidrule(l){2-8}
      & Test Set~2 & 0.205 & 0.237 & 0.224 & 0.729 & 0.662 & 0.653 \\
    \midrule
    \multirow[t]{2}{*}{Model E}
      & Test Set~1 & 0.130 & 0.150 & 0.151 & 0.983 & 0.978 & 0.978 \\
    \cmidrule(l){2-8}
      & Test Set~2 & 0.143 & 0.166 & 0.147 & 0.868 & 0.835 & 0.850 \\
    \midrule
    \multirow[t]{2}{*}{Model F}
      & Test Set~1 & 0.197 & 0.219 & 0.217 & 0.961 & 0.952 & 0.955 \\
    \cmidrule(l){2-8}
      & Test Set~2 & 0.132 & 0.145 & 0.128 & 0.887 & 0.872 & 0.887 \\
    \bottomrule
  \end{tabular}
  
\end{table}

\section{Discussion}
\label{subsec4}
As mentioned in the previous section, the shallower models, Model A and Model B, showed a significant decline in performance on Test Set 2 as shown in  \Cref{tab:test_performance}. Although Model C was selected based on its slightly better validation performance and training stability, deeper configurations such as Model E and Model F achieved higher performance on Test Set 2. \Cref{tab:scatter} illustrates the scatter plots of measured versus predicted illuminance values for Model C, Model E, and Model F on both Test Set 1 and Test Set 2, providing further insight into their systematic bias and generalization abilities.

On Test Set 1, all three models (C, E, and F) exhibited regression lines that closely followed the identity line (\(y = x\)), indicating minimal prediction bias. The data points were tightly clustered around the ideal line, suggesting high predictive accuracy and consistent performance across all three illuminance indicators.

In contrast, on Test Set 2, Model E (dropout = 0.3) showed a slightly higher \(R^2\) than Model C, but exhibited the largest deviation from the identity line, with a clear pattern of underestimating high values and overestimating low values. This indicates a systematic prediction bias, suggesting that the model fails to capture the full dynamic range of the target variable. On the other hand, Model F (dropout = 0.5) achieved the highest \(R^2\) value on Test Set 2, with all three illuminance indicators (\(E_h\), \(E_s\), and \(E_e\)) showing \(R^2\) values greater than 0.87. As shown in \Cref{tab:scatter}, the regression lines of Model F on Test Set 2 were closely aligned with the identity line, similarly to Model C, indicating a comparable degree of prediction consistency.

These results indicate that a well-balanced combination of network capacity and regularization is crucial for achieving robust generalization under changing temporal and environmental conditions. Therefore, future work could explore deeper and wider network architectures, combined with appropriate regularization techniques, to improve model generalization under temporal and weather-related variations.

{
  
  \renewcommand{\baselinestretch}{1.0}\selectfont  
  \scriptsize                                      
  \renewcommand{\arraystretch}{1.2}                
  \setlength{\tabcolsep}{4pt}

  \setlength{\LTcapwidth}{\textwidth}

  \begin{longtable}[l]{@{} 
      >{\raggedright\arraybackslash}p{1.5cm}   
      >{\raggedright\arraybackslash}p{1.8cm}   
      >{\raggedright\arraybackslash}p{2.5cm}   
      >{\raggedright\arraybackslash}p{2.5cm}   
      >{\raggedright\arraybackslash}p{2.5cm}   
  @{}}

    \caption{The scatter plots of observed and predicted values on Test Set 1 and Test Set 2.} \label{tab:scatter} \\

    \toprule
    Model & Test Set & $E_h$ & $E_s$ & $E_e$ \\
    \midrule
    \endfirsthead

    \multicolumn{5}{@{}l}{{\scriptsize \tablename\ \thetable{} -- Continued from previous page}} \\
    \toprule
    Model & Test Set & $E_h$ & $E_s$ & $E_e$ \\
    \midrule
    \endhead

    \midrule
    \multicolumn{5}{r}{{\scriptsize Continued on next page}} \\
    \endfoot

    \bottomrule
    \endlastfoot
    
    Model C 
      & Test Set 1 & \multicolumn{3}{l}{\raisebox{-\height}{\includegraphics[width=0.8\linewidth]{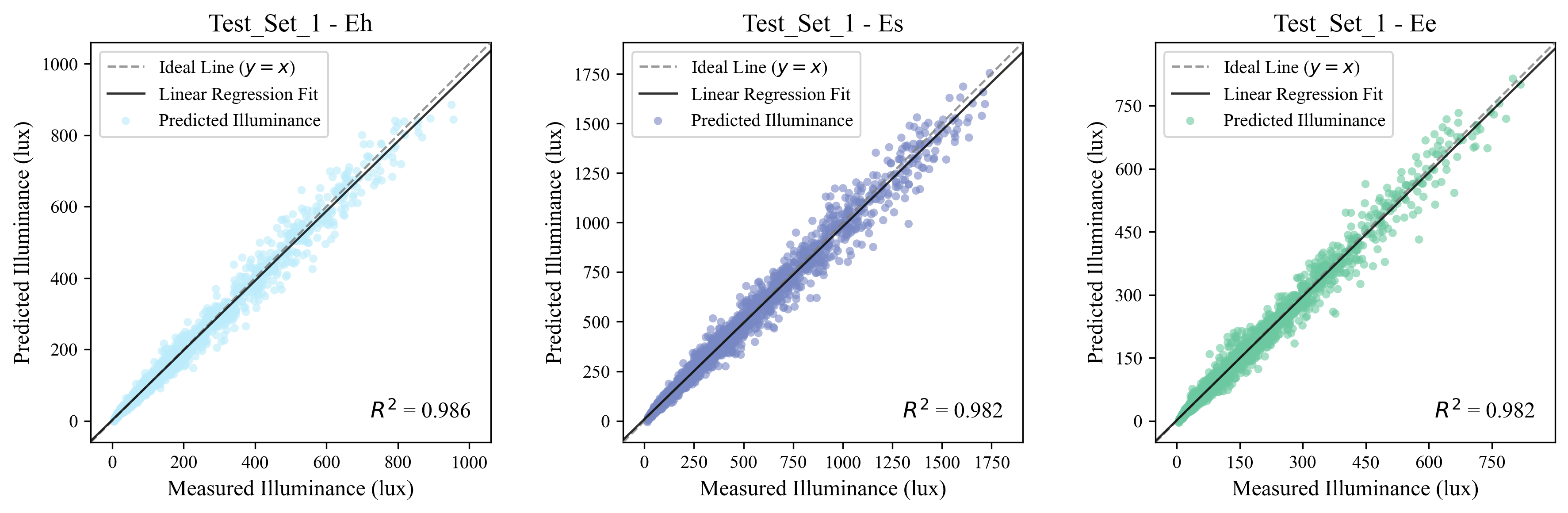}}} \\
      \addlinespace[10pt]
      & Test Set 2 & \multicolumn{3}{l}{\raisebox{-\height}{\includegraphics[width=0.8\linewidth]{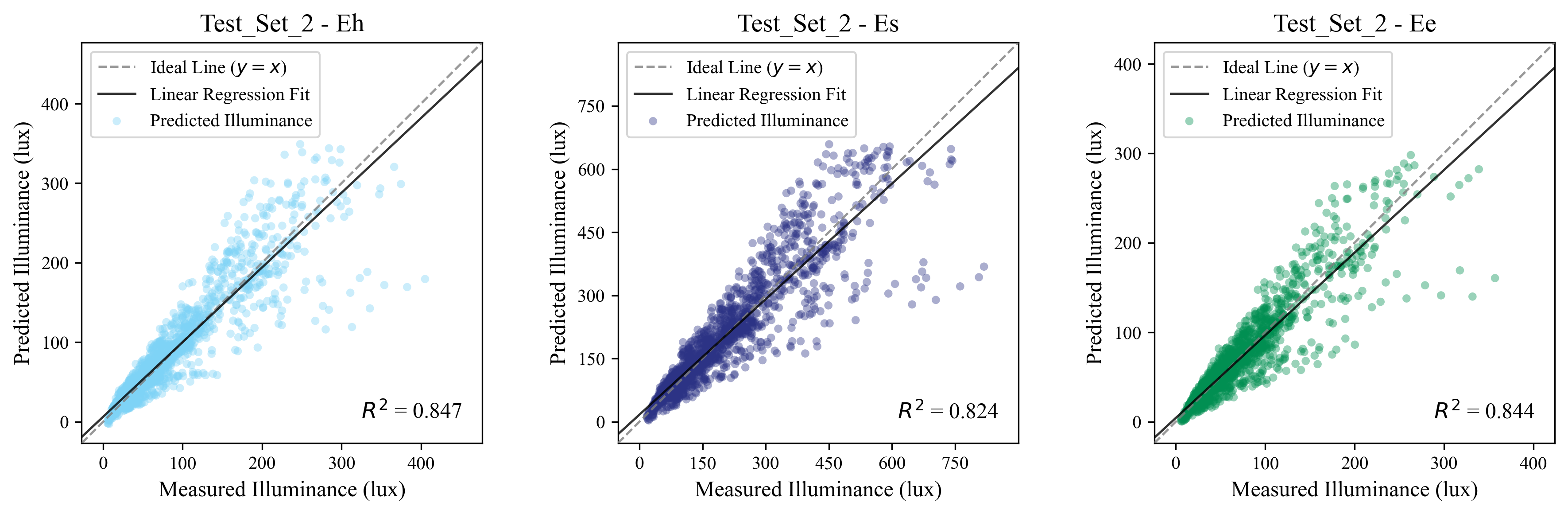}}} \\
    
    \cmidrule(lr){2-5} \addlinespace[5pt]
    
    Model E
      & Test Set 1 & \multicolumn{3}{l}{\raisebox{-\height}{\includegraphics[width=0.8\linewidth]{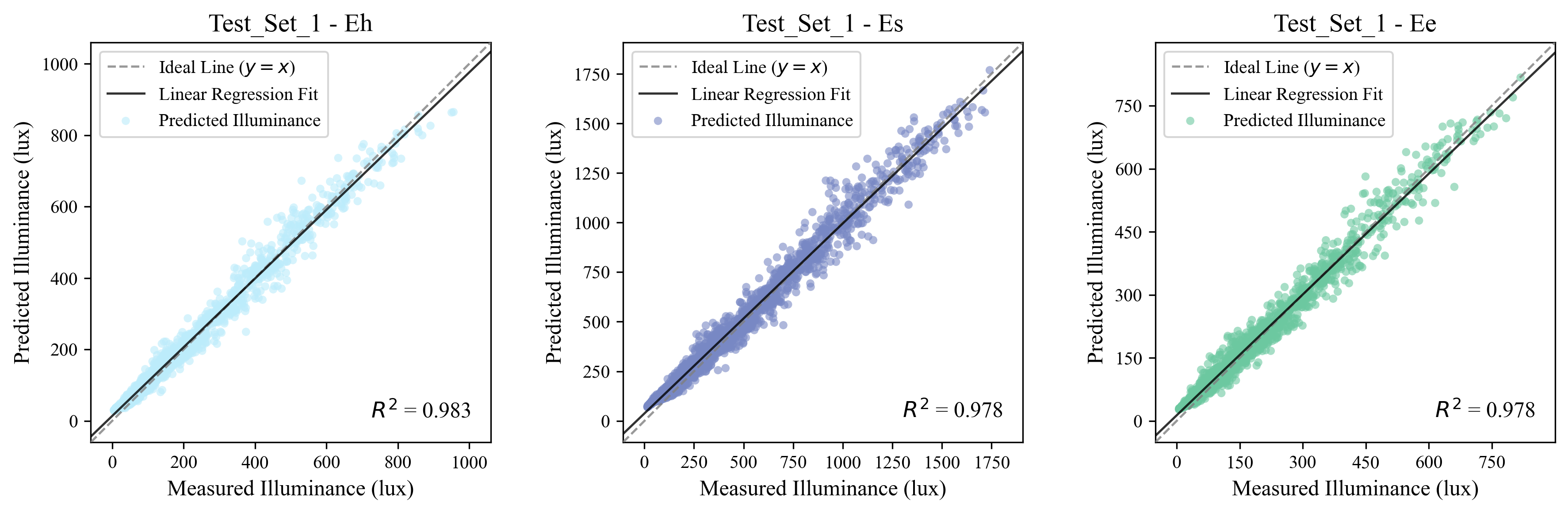}}} \\
      \addlinespace[10pt]
      & Test Set 2 & \multicolumn{3}{l}{\raisebox{-\height}{\includegraphics[width=0.8\linewidth]{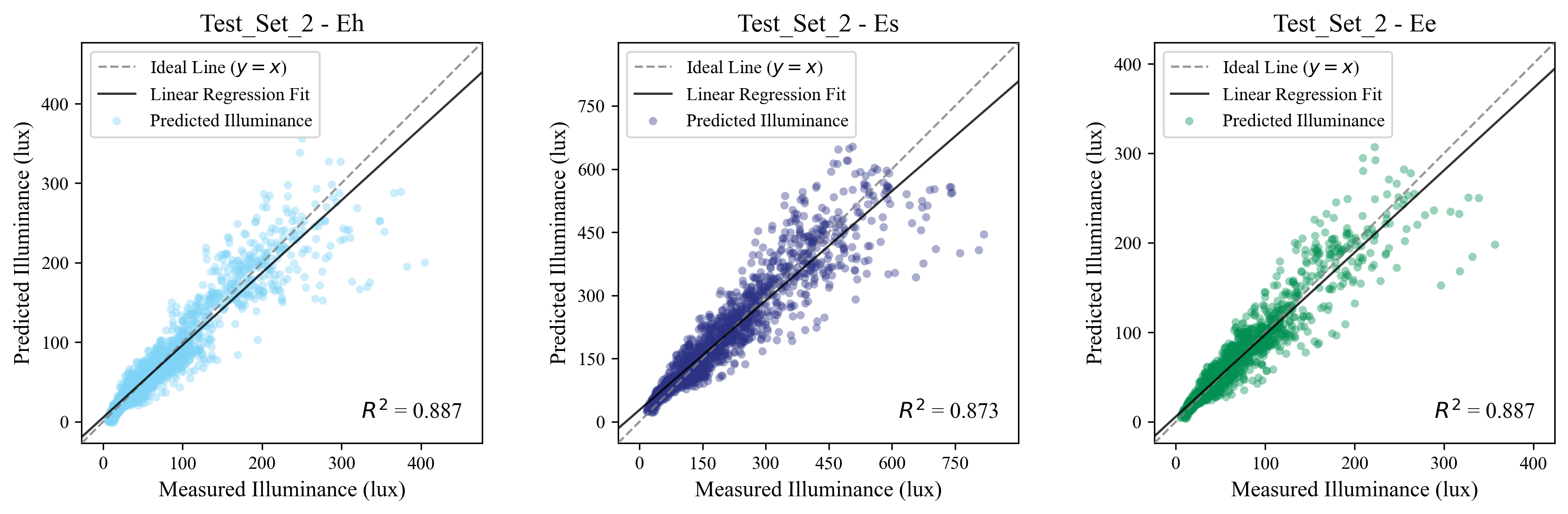}}} \\
    
    \cmidrule(lr){2-5} \addlinespace[5pt]
    
    Model F
      & Test Set 1 & \multicolumn{3}{l}{\raisebox{-\height}{\includegraphics[width=0.8\linewidth]{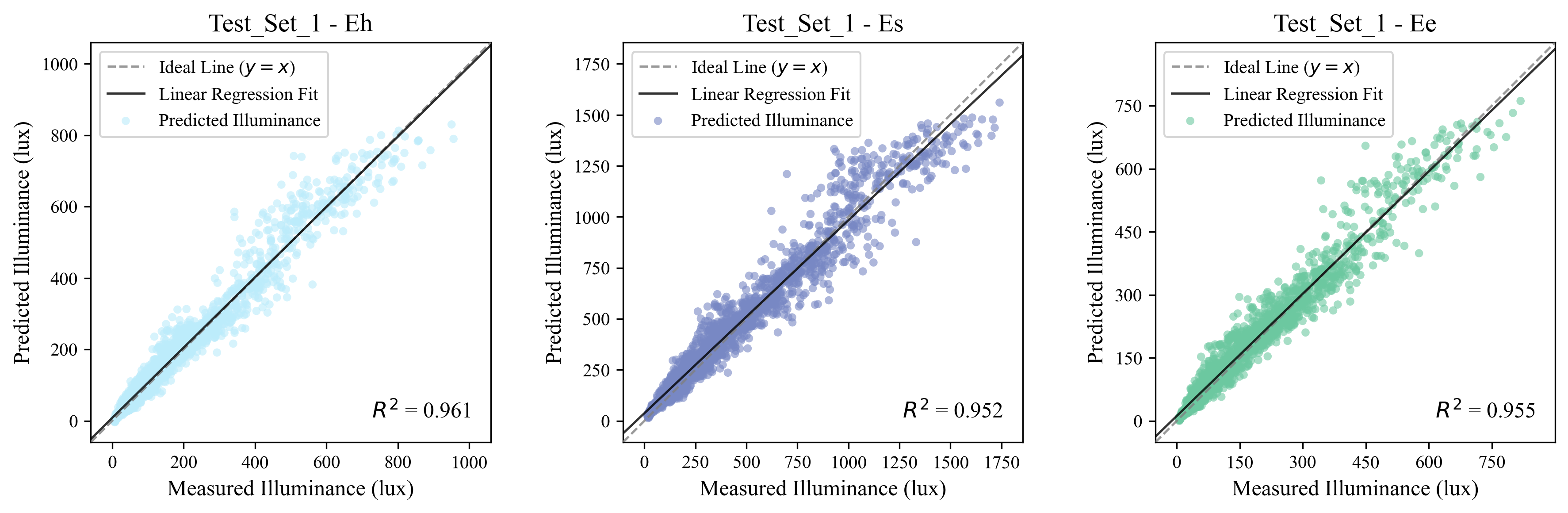}}} \\
      \addlinespace[10pt]
      & Test Set 2 & \multicolumn{3}{l}{\raisebox{-\height}{\includegraphics[width=0.8\linewidth]{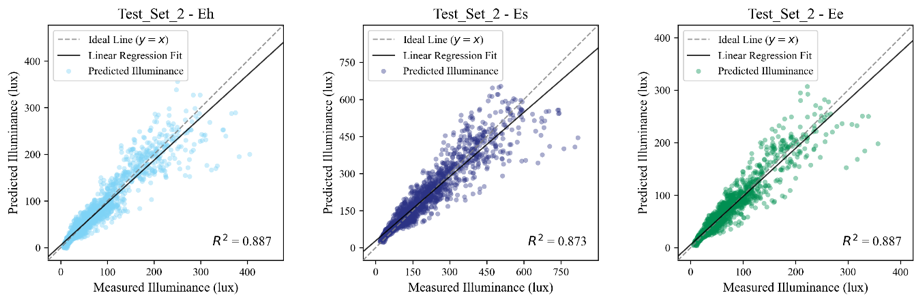}}} \\

  \end{longtable}
}
To assess the model’s accuracy under same-distribution conditions, eight representative samples were selected from Test Set~1. Table~\ref{tab:case_study} presents the measured and predicted illuminance values across three target illuminance values (\(E_h\), \(E_s\), \(E_e\)) of these samples, along with their corresponding preprocessed, non-intrusive window-domain images, spatial and temporal features, and maximum absolute errors. These samples cover a diverse range of features, including different times of day and various sensor locations, from near-window to deep-room positions.

As shown in Table~\ref{tab:case_study}, most samples exhibit maximum absolute errors below 60\,lux, indicating that the model performs well under same-distribution conditions. This further verifies the effectiveness of the proposed non-intrusive, image-based illuminance prediction framework, which can effectively capture the key visual features from image inputs and spatial and temporal features. Most maximum errors occurred in \(E_s\), which might be attributed to the highest illuminance values among the three illuminance indicators. In high-illuminance samples such as Case~1, Case~2, Case~3, and Case~5, the corresponding maximum errors exceed 50\,lux, while low-illuminance cases such as Case~4, Case~6, and Case~7 consistently yield smaller errors (all \(<20.0\)\,lux). This indicates that the model achieves better accuracy in lower-illuminance areas. These results may be explained by the data distribution illustrated in ~\Cref{fig:illuminance_dist}, which shows that the dataset consists primarily of low-illuminance samples, with relatively fewer high-illuminance instances.

Therefore, to improve generalization in high-illuminance environments such as areas near windows with abundant daylight, future work should focus on collecting and integrating more high-illuminance training samples to enhance the diversity of the dataset.

{
  \renewcommand{\baselinestretch}{1.0}\selectfont  
  \scriptsize                                      
  \renewcommand{\arraystretch}{1.2}                
  \setlength{\tabcolsep}{4pt}

  \begin{longtable}{@{} 
      >{\raggedright\arraybackslash}p{0.6cm}   
      >{\raggedright\arraybackslash}p{3.0cm}   
      >{\raggedright\arraybackslash}p{1.6cm}   
      >{\raggedright\arraybackslash}p{0.8cm}   
      >{\raggedright\arraybackslash}p{2.8cm}   
      >{\raggedright\arraybackslash}p{2.8cm}   
      >{\raggedright\arraybackslash}p{1.0cm}   
  @{}}
    
    \caption{Measured and predicted illuminance results for eight selected samples from Test Set~1.} \label{tab:case_study} \\

    \toprule
    Case & Image & Sensor ID (X, D) & Time 
         & Measured \newline [\(E_h\), \(E_s\), \(E_e\)] (lux) 
         & Predicted \newline [\(E_h\), \(E_s\), \(E_e\)] (lux) 
         & Error \\
    \midrule
    \endfirsthead

    \multicolumn{7}{l}{{\scriptsize \tablename\ \thetable{} -- Continued from previous page}} \\ 
    \toprule
    Case & Image & Sensor ID (X, D) & Time 
         & Measured \newline [\(E_h\), \(E_s\), \(E_e\)] 
         & Predicted \newline [\(E_h\), \(E_s\), \(E_e\)] 
         & Error \\
    \midrule
    \endhead

    \midrule
    \multicolumn{7}{r}{{\scriptsize Continued on next page}} \\
    \endfoot

    \bottomrule
    \endlastfoot
    
    1 & \includegraphics[width=\linewidth,height=1.2cm,keepaspectratio]{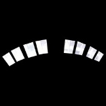}
      & (5.8, 1.5) & 8:40
      & [374.8, 700.4, 141.0]
      & [305.2, 570.1, 126.7]
      & 130.3 \\

    2 & \includegraphics[width=\linewidth,height=1.2cm,keepaspectratio]{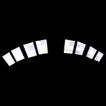}
      & (5.8, 1.5) & 16:00
      & [217.9, 404.4, 77.9]
      & [198.0, 346.3, 56.0]
      & 58.1 \\

    3 & \includegraphics[width=\linewidth,height=1.2cm,keepaspectratio]{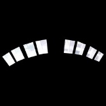}
      & (4.3, 1.5) & 13:45
      & [588.8, 1330.2, 575.8]
      & [559.1, 1251.3, 568.8]
      & 78.9 \\

    4 & \includegraphics[width=\linewidth,height=1.2cm,keepaspectratio]{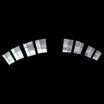}
      & (2.8, 1.5) & 11:45
      & [36.3, 89.3, 37.5]
      & [32.5, 78.4, 31.1]
      & 10.9 \\

    5 & \includegraphics[width=\linewidth,height=1.2cm,keepaspectratio]{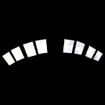}
      & (2.8, 3.0) & 14:25
      & [371.5, 843.7, 301.6]
      & [403.1, 897.7, 327.0]
      & 54.0 \\

    6 & \includegraphics[width=\linewidth,height=1.2cm,keepaspectratio]{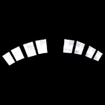}
      & (5.8, 4.5) & 8:50
      & [74.2, 209.1, 63.8]
      & [64.5, 225.2, 64.0]
      & 16.0 \\

    7 & \includegraphics[width=\linewidth,height=1.2cm,keepaspectratio]{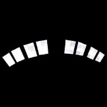}
      & (5.8, 6.0) & 9:05
      & [33.9, 87.8, 30.5]
      & [40.0, 101.3, 34.6]
      & 13.5 \\

    8 & \includegraphics[width=\linewidth,height=1.2cm,keepaspectratio]{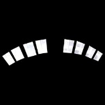}
      & (5.8, 6.0) & 15:05
      & [66.0, 175.7, 64.6]
      & [79.6, 205.4, 74.5]
      & 29.6 \\

  \end{longtable}
}

\begin{figure}[htbp!]
  \centering
  \includegraphics[width=\textwidth, height=0.3\textheight, keepaspectratio]{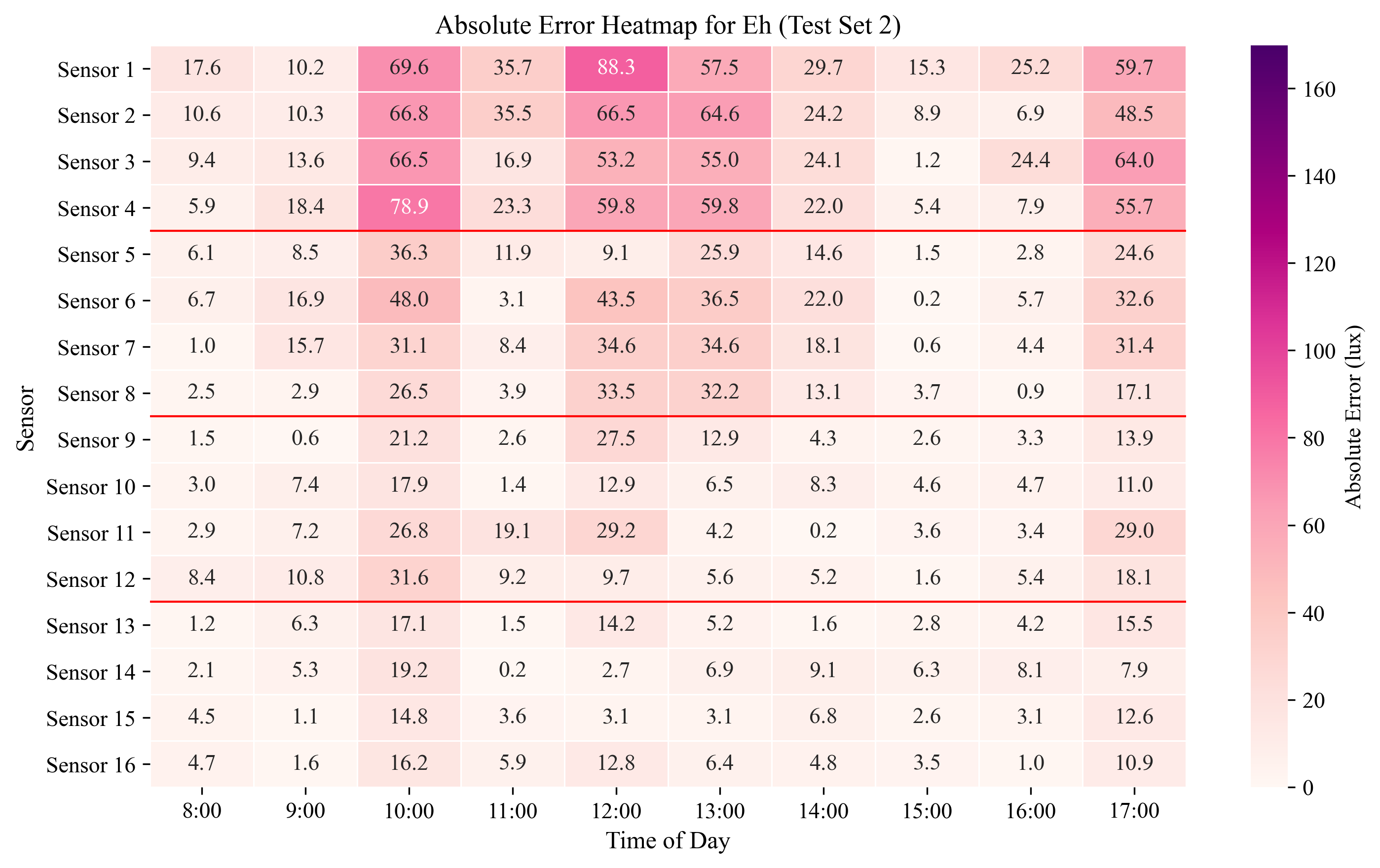}\\[1ex]
  \includegraphics[width=\textwidth, height=0.3\textheight, keepaspectratio]{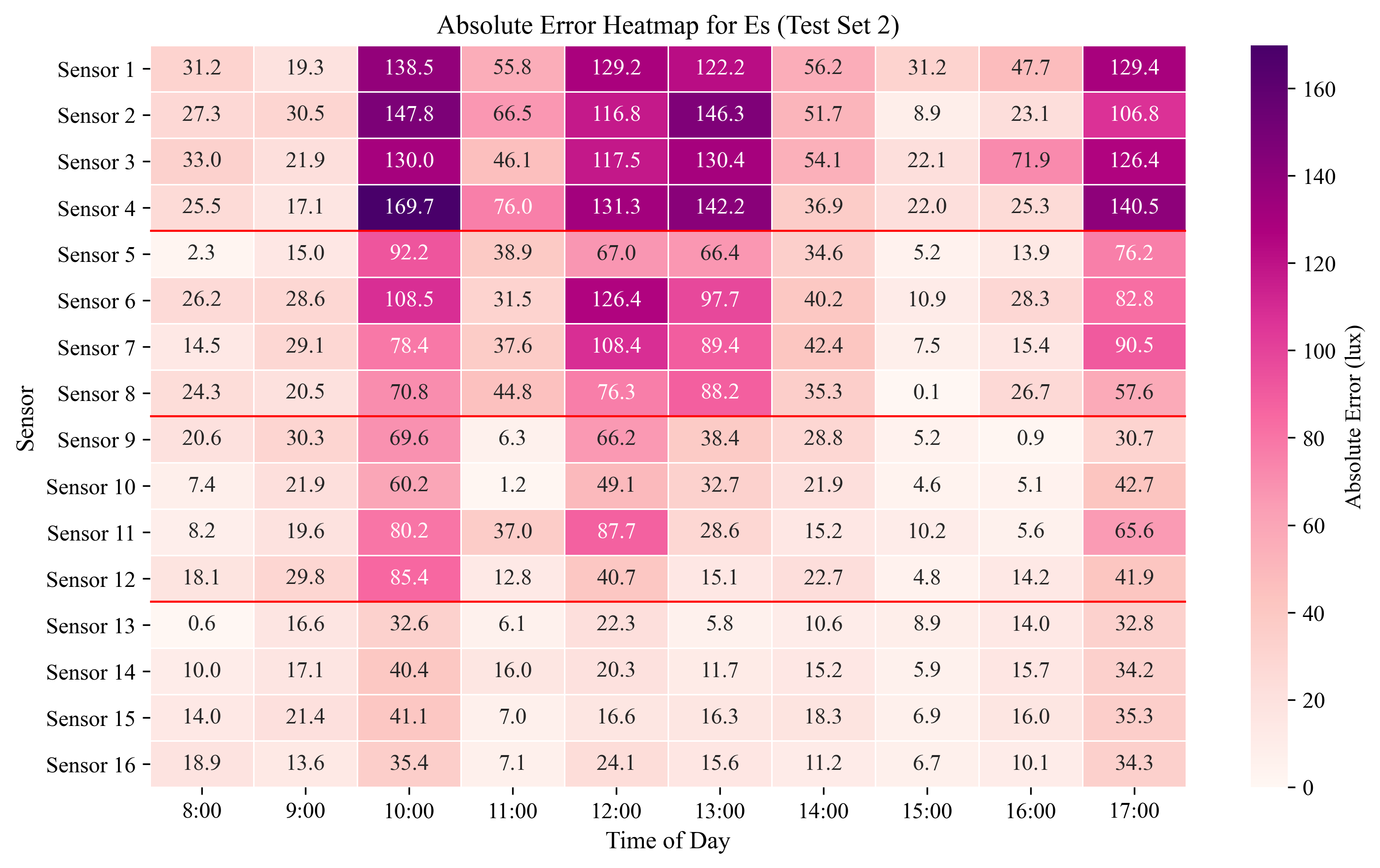}\\[1ex]
  \includegraphics[width=\textwidth, height=0.3\textheight, keepaspectratio]{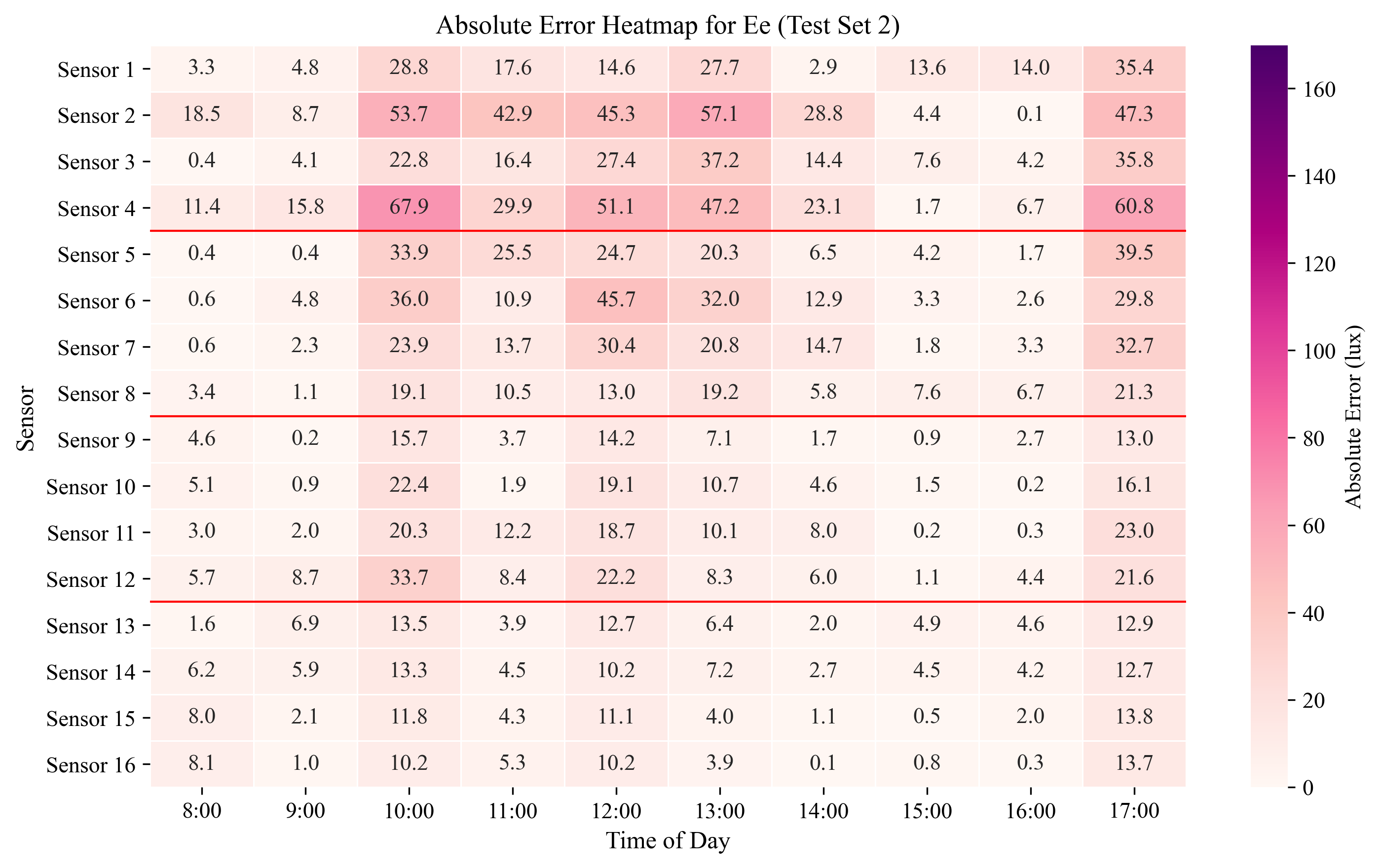}
  \caption{Heatmaps of prediction error distributions on Test Set~2.}
  \label{fig:error_heatmaps}
\end{figure}

Additionally, the generalization ability of the model on new data is noteworthy. In this study, to evaluate the model’s performance on temporally unseen scenarios, we further discussed the model’s predictions on Test Set~2, which contains measurement data from a future day not in the training phase. The absolute error heatmaps of predicted illuminance for \(E_h\), \(E_s\), and \(E_e\) on Test Set~2 are illustrated in ~\cref{fig:error_heatmaps}. The heatmaps visualize the spatio–temporal distribution of prediction errors across 16 sensors and 10 hourly time points. Red horizontal lines are used to separate sensors by rows according to their spatial arrangement from near-window (Sensors~1–4) to deep-room positions (Sensors~13–16).

Overall, the highest prediction errors are concentrated in the near-window area (Sensors~1–4) during midday (10:00–13:00) and again at 17:00, especially for the \(E_s\) indicator, with the maximum error exceeding 160\,lux. In addition, the error tends to decrease with increasing distance from the window. The lowest errors are observed in the deepest row (Sensors~13–16), where most absolute errors of \(E_e\) and \(E_h\) are lower than 10\,lux. This trend is likely due to the more stable and diffuse lighting conditions in deeper zones of the room, where daylight variation is less pronounced and less affected by direct sunlight.

These results suggest that while the model demonstrates robust performance in stable and low-illuminance areas, its accuracy tends to decline under high-illuminance, near-window conditions. This may be attributed to three main factors. First, near-window areas are prone to more intense and dynamic daylight at certain times, including direct sunlight and moving shadows. These factors increase visual complexity and may limit the model’s ability to extract effective features from a single image. Second, high-value samples constitute the majority in the training dataset, as mentioned in the previous section. This data scarcity leads to a decrease in the model’s generalization ability in high-illuminance scenarios. Third, absolute errors tend to be larger in high-illuminance scenarios even under comparable relative error rates.

\begin{figure}[htbp!]
  \centering
  \includegraphics[width=\linewidth,keepaspectratio]{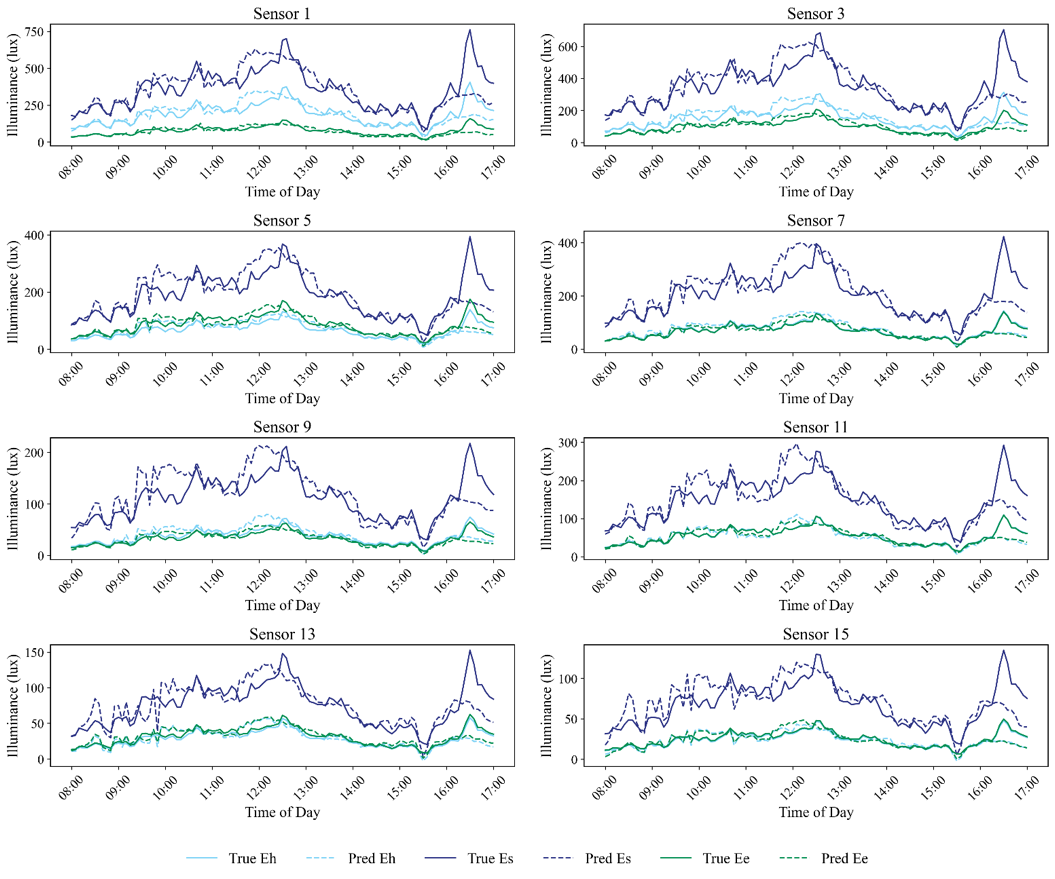}
  \caption{The measured and predicted illuminance values of Eh, Es, and Ee for selected sensors across varying room depth on an unseen day (Test Set 2).}
  \label{fig:time_series}
\end{figure}

~\cref{fig:time_series} presents time‐series plots of the measured and predicted illuminance values (\(E_h\), \(E_s\), and \(E_e\)) across a full unseen day for eight selected sensors located at varying distances from the window. The predicted illuminance curves exhibit good alignment with the measured curves, indicating the model’s ability to effectively capture daylight dynamics and validating its temporal robustness under unseen lighting conditions.

However, between approximately 09:30–13:00 and 16:00–17:00, the predicted curves of the \(E_s\) indicator show noticeable divergence from the measured curves, particularly in near‐window sensors such as Sensor~1 and Sensor~3. This observation is consistent with the error patterns observed in ~\cref{fig:error_heatmaps}. Additionally, the sharp changes in the measured illuminance during these periods further support the explanation of error patterns proposed in Fig.~\ref{fig:error_heatmaps}: that the prediction errors are primarily due to dynamic daylight conditions. Furthermore, these deviations are likely driven by the same factors discussed in ~\cref{fig:error_heatmaps}, including rapid daylight changes in near‐window areas, reduced feature reliability under dynamic conditions, and the limited representation of high‐illuminance samples in the training dataset.

In this study, we propose a non-intrusive, image-based approach to predict the indoor daylight environment, utilizing a ceiling-mounted camera to capture the image features of side-lit window areas, without installing any sensors in the occupied space. The predicted model can serve as a foundational input module of DLCs, thereby further enabling energy savings~\cite{kim2022performance}. By integrating it into a Building Management System (BMS) or the Internet of Things (IoT) lighting platform, control algorithms can dynamically adjust LED output and motorized blinds to maintain target light levels in real-time while minimizing energy consumption. For instance, when natural daylight is abundant, lamp power is automatically reduced, while when daylight is scarce, artificial lighting is precisely supplemented only as needed. Furthermore, the method is low-cost, easy to maintain, and able to rapidly predict accurate results, providing a flexible and reliable solution for future DLC implementation in complex dynamic environments to achieve energy savings.  

While the proposed framework shows promising results on both same-distribution and unseen datasets, there are several limitations that should be noted, which may be addressed in future work. In this study, the model architecture and hyperparameters were tuned using the validation dataset, achieving good performance on the same-distribution test dataset (Test Set~1). However, performance dropped on temporally unseen data (Test Set~2). Future research could expand the dataset and explore deeper and wider architectures. For example, Model~F in Table~\ref{tab:scatter}, which employs a greater number of hidden nodes, demonstrated better performance on Test Set~2.

Additionally, the model’s performance tends to decline in high-illuminance scenarios, likely due to the data scarcity of such samples during model training. Future work could expand the dataset with additional high-luminance samples and explore data augmentation techniques to enrich the dataset under diverse daylighting conditions. Besides, the prediction accuracy for \(E_s\) values is generally lower than for \(E_h\) and \(E_e\). This may be caused by the limited ability of the single image to capture the direct daylight features. Thus, another important direction is enhancing the model's ability to capture directional daylight features, possibly through multi-view imaging and additional spatial inputs. Moreover, the proposed approach can serve as a foundation for rapidly providing real-time inputs to daylighting control systems, with potential applications in smart lighting control and energy-efficient building management. However, the current research remains at the prediction stage.

\section{Conclusion}
\label{subsec5}
This study develops a non-intrusive multimodal deep learning framework for real-time prediction of the distribution of indoor workplane illuminance from three directions (\(E_h\), \(E_s\), \(E_e\)). By integrating images of side-lit window regions captured by a ceiling-mounted camera with structured data such as temporal and spatial features, the proposed CNN-MLP model effectively captures complex dynamic daylight patterns, even under scenarios with indoor activities. A field experiment was conducted in a test room in Guangzhou, China, where a dataset of 17,344 samples was collected and split into a training set and two test sets, designed respectively for same-distribution validation and new-day generalization. Results demonstrate that the selected model achieved high predictive accuracy ($R^2 > 0.98$ on Test Set~1, with the same-distribution) and maintained a certain generalization to unseen scenarios ($R^2 > 0.82$ on Test Set~2, temporally unseen), showing the potential of multimodal deep learning to capture diverse and complex features for accurate daylight prediction.

The proposed method assumes the feasibility of using low-cost LDR images for accurate daylight prediction in dynamically changing indoor environments. However, prediction errors of the model increased in high-illuminance scenarios. These errors are tolerable for practical applications of lighting control, as such scenarios typically receive sufficient daylight and therefore require little supplemental electric lighting. In addition, the dataset was collected in a single test room, which may limit the generalization of the model to other climates and building types. Besides, the use of single-view LDR images, while ensuring simplicity and low cost, limits the range of daylighting informations that can be captured. Nevertheless, this study primarily focuses on the real-time data acquisition of indoor daylighting information as a foundation for DLCs, while further issues such as control strategies, occupant comfort, and system integration remain to be explored in future applications.

Future research could expand the dataset across diverse scenarios to enhance the model's generalization performance. Besides, multi-view and HDR techniques could be considered to better capture high-luminance window features. Moreover, the framework of this study serves as a foundation for DLCs, upon which future work will further investigate control strategies and energy-saving performance.

\section*{CRediT authorship contribution statement}
\textbf{Zulin Zhuang:} Writing – original draft, Software, Methodology, Data curation, Visualization. 
\textbf{Yu Bian:} Writing – review \& editing, Supervision, Funding acquisition, Methodology, Conceptualization.

\section*{Declaration of competing interest}
The authors declare that they have no known competing financial interests or personal relationships that could have appeared to influence the work reported in this paper.

\section*{Data availability}
The data that has been used is confidential.

\section*{Acknowledgments}
This research was supported by the National Natural Science Foundation of China (No. 52278107), the Natural Science Foundation of Guangdong Province (No. 2024A1515011428), and State Key Laboratory of Subtropical Building and Urban Science (No. 2024ZB04).

  \bibliographystyle{elsarticle-num} 
  \bibliography{references}
  
\end{document}